\documentclass[11pt]{article}

\usepackage[final]{acl}
\usepackage{booktabs}
\usepackage[most]{tcolorbox} 
\usepackage{xcolor}
\usepackage{times}
\usepackage{latexsym}
\usepackage{mathtools}
\usepackage{multirow}
\usepackage{amsfonts}
\usepackage{enumitem}
\usepackage{amsmath}
\usepackage{amssymb}
\usepackage{amsthm}
\usepackage[T1]{fontenc}

\usepackage[utf8]{inputenc}

\usepackage{microtype}
\usepackage{algorithm}
\usepackage{algorithmic}
\usepackage{inconsolata}

\usepackage{graphicx}
\newtcolorbox{promptbox}[1]{
  enhanced,
  colback=white,
  colframe=black,
  colbacktitle=black,
  coltitle=white,
  fonttitle=\bfseries\small,
  arc=8pt,
  boxrule=1pt,
  left=10pt, right=10pt, top=10pt, bottom=10pt,
  title=#1,
  attach boxed title to top left={xshift=0mm, yshift=0mm},
  boxed title style={sharp corners, boxrule=0pt}
}
\newtheorem{proposition}{Proposition}
\newcommand{\var}[1]{\textcolor{brown}{\texttt{\{#1\}}}}
\title{Graph-Enhanced Policy Optimization in LLM Agent Training}

\author{Jiazhen Yuan, Zhike Gong, Jinquan Hang, Zhengbiao Bai, and Wei Zhao \\
  JD.com \\
  \texttt{\{yuanjiazhen.jason, gongzhike.1, hangjinquan1, baizhengbiao1, zhaowei87\} @jd.com}
}
\begin{document}
\maketitle
\begin{abstract}

Multi-step LLM agents in interactive environments represent a crucial step toward long-horizon decision-making. 
To train such agents, group-based reinforcement learning is widely adopted, which reinforces trajectories with higher relative performance within the group. 
However, in most existing methods, every step within a trajectory and every trajectory with the same terminal reward receive identical credit, regardless of their actual contributions. 
Since different states play different structural roles in an online state-transition graph built from sampled trajectories, their impacts should be differentiated and converted into task-aware credit at both the step and trajectory levels. 
We therefore present Graph-Enhanced Policy Optimization (GEPO), a framework for dual-level structural credit assignment in multi-step LLM agent training. 
Specifically, GEPO derives a state-level Task-Conditioned Criticality score that combines topological betweenness on the state-transition graph with semantic similarity to the task prompt. 
Based on this score, trajectory-level credit is reshaped through a state-adaptive discount, while step-level credit is scaled by the criticality of its successor state. 
Experimental results show that GEPO outperforms the strongest baselines by 1.1\% in success rate on ALFWorld, 3.2\% on WebShop, and 3.8\% on average across search-augmented QA tasks at the 7B scale. Compared with flat group-based methods, GEPO reduces across-seed variance and concentrates gradient signals on the most critical steps.

\end{abstract}
\section{Introduction}

\begin{figure}[t]
    \centering
    \includegraphics[width=\columnwidth, trim={350pt 0pt 450pt 0pt}, clip]{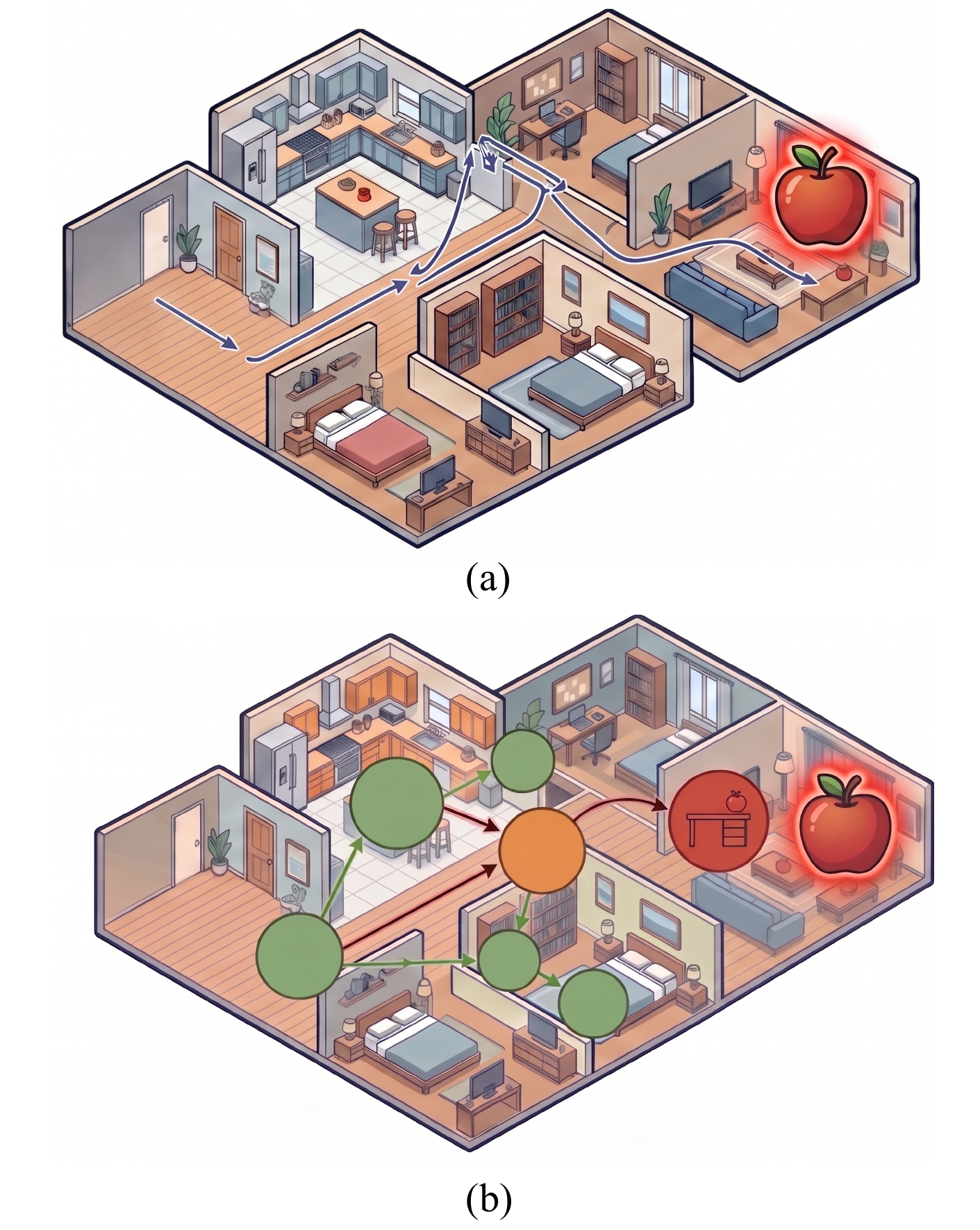}
    \caption{Comparison of credit assignment mechanisms. \textbf{(a)} Standard group-based methods suffer from \textbf{structural blindness}, uniformly distributing sparse rewards across all steps and causing gradient noise. \textbf{(b)} GEPO constructs an online topological graph to identify bottleneck states. This allows for redistributing the advantage precisely to the most critical step.}
    \label{fig:introduction}
\end{figure}

Multi-step LLM agents~\cite{Yao2022-xt} in interactive environments represent a crucial step toward long-horizon decision-making.
These agents~\cite{Chen2024-gf} have been widely studied in tasks such as virtual home navigation~\cite{Shridhar2020-fd}, multi-step web shopping~\cite{Yao2022-yg, Furuta2023-re}, and game-like world exploration~\cite{Wang2024-voyager}.
The dominant approach for training such agents~\cite{Quadros2025-vs} is group-based reinforcement learning~\cite{Shao2024-jf}, which compares trajectories sampled from the same prompt and reinforces those with higher relative performance.
However, in most existing methods, every step within a trajectory and every trajectory with the same terminal reward receive identical credit, regardless of their actual contributions~\cite{Islam2022-sx}.
In ALFWorld~\cite{Shridhar2020-fd}, for example, an agent must clear bottleneck states such as room transitions and container transitions before any reward arrives, yet these pivotal steps receive no more credit than routine ones.
This \textbf{structural blindness} inflates the variance of policy updates~\cite{Ying2024-ab} and slows learning in topologically complex environments~\cite{Ramakrishnan2018-ex}.

The underlying topology of such environments can be recovered online by building a state-transition graph from each batch of sampled trajectories~\cite{Hazra2023-kp}.
On this graph, individual steps and entire trajectories can be differentiated by the topological role of the states they visit, such as their betweenness centrality on the graph.
This per-state structural signal can then be translated into task-aware credit at both the trajectory and step level.

Building on this idea, we introduce \textbf{Graph-Enhanced Policy Optimization (GEPO)}, a framework that performs dual-level structural credit assignment for LLM agent training.
Each state receives a Task-Conditioned Criticality score that combines its betweenness centrality on the graph with its semantic similarity to the task prompt, so that states which are both structurally pivotal and task-relevant receive the highest scores.
At the trajectory level, this score reshapes the return through a state-adaptive discount factor, allowing trajectories that reach the same reward via structurally different paths to receive distinguishable advantages.
At the step level, the resulting trajectory advantage is redistributed onto individual actions and scaled by the successor state's criticality, so that steps leading to critical states are reinforced more strongly than those leading to routine ones.

Our contributions are as follows:

\begin{itemize}[leftmargin=*]
\item To the best of our knowledge, we are among the first to introduce structural credit assignment through an online state-transition graph for LLM agent training, which adaptively reinforces critical steps and trajectories.

\item We propose GEPO, a training framework that performs dual-level credit assignment at both the trajectory and step level using the state-transition graph. Each state on the graph receives a \textbf{Task-Conditioned Criticality} score that fuses betweenness centrality with semantic similarity to the task prompt. Based on this score, the trajectory's credit is reshaped through a state-adaptive discount, and each step's credit is scaled by its successor state's criticality.

\item GEPO outperforms the strongest baselines by 1.1\% in success rate on ALFWorld, 3.2\% on WebShop~\cite{Yao2022-yg}, and 3.8\% on average across search-augmented QA tasks at the 7B scale. Compared with flat group-based methods, GEPO reduces across-seed variance and concentrates gradient signals on the most critical steps.
\end{itemize}

\begin{figure*}[t]
    \centering
    \hspace*{0.3cm}
    \includegraphics[width=\textwidth, trim={0pt 100pt 0pt 20pt}, clip]{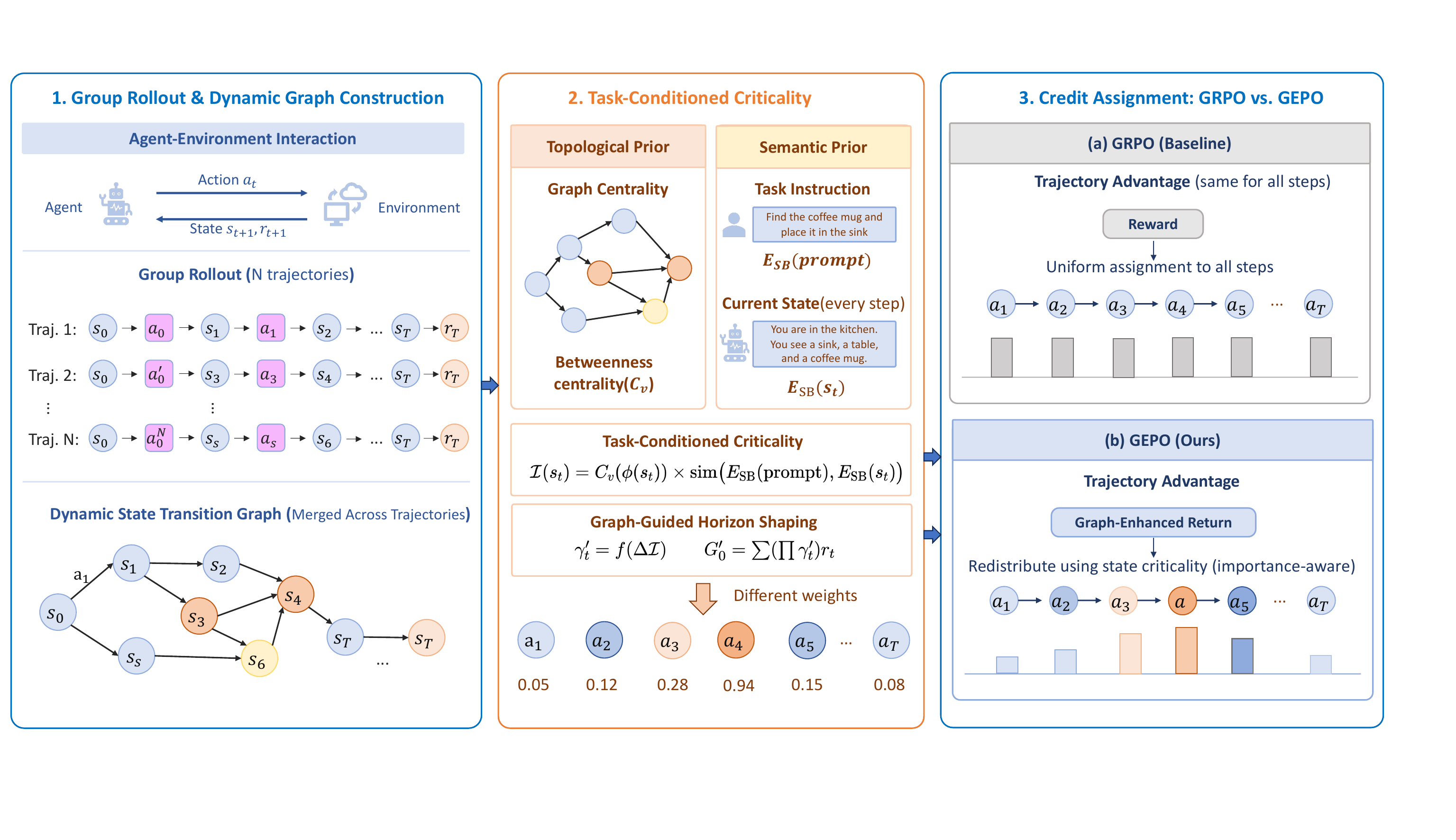}
    \caption{An overview of the GEPO framework. \textbf{(1)} The agent collects a group of trajectories through environment interaction, which are merged to construct a dynamic state-transition graph. \textbf{(2)} Task-Conditioned Criticality is computed by fusing betweenness centrality and semantic similarity, guiding the dynamic horizon shaping. \textbf{(3)} Unlike standard GRPO, which assigns uniform advantages across all steps, GEPO redistributes the trajectory-level advantage to individual steps based on their graph-derived criticality, enabling precise credit assignment.}
    \label{fig:gepo_framework}
\end{figure*}

\section{Related Work}
\label{sec:related}

Recent years have witnessed a rapid surge in efforts to develop LLM-based agents for complex multi-step tasks~\cite{Luo2025-sa}. Beyond single-step Q\&A or text generation, these agents integrate large language models with environment interactions, enabling them to navigate virtual homes~\cite{Shridhar2020-fd}, browse the web for multiple steps~\cite{Yao2022-yg}, and solve puzzle-like problems~\cite{Wang2024-xi}. While such agents have demonstrated promising capabilities~\cite{Chen2024-gf}, they often face significant reward sparsity and partially observable states~\cite{Quadros2025-vs}, making reinforcement learning a natural choice for enabling long-horizon control~\cite{Erdogan2025-cb}.

A number of policy optimization methods have been adapted to LLMs, including PPO~\cite{Schulman2017-dh} and DPO~\cite{Rafailov2023-eu}. However, these typically rely on dense rewards or train computationally expensive critic networks. Recent work has turned to group-based RL methods that learn from sparse feedback without a value network~\cite{Ouyang2022-xd,Sharma2024-fv,Lin2025-gu}. For instance, GRPO~\cite{Shao2024-jf} compares trajectories within a group sharing the same prompt. While memory-efficient, GRPO distributes terminal rewards uniformly across all steps, exhibiting the \textbf{structural blindness} that injects high-variance gradient noise. To refine credit assignment, GiGPO~\cite{Feng2025-oa} attempts empirical state matching across trajectories and uses the matched states directly as the unit of credit attribution. However, in combinatorially large textual environments, exact state overlaps are virtually nonexistent, and using approximate matches as the credit unit propagates matching errors directly into the step-level advantage. Similarly, recent extensions like HGPO~\cite{He2026-qk} attempt to resolve historical context inconsistency through hierarchical history matching, which shares GiGPO's vulnerability to matching sparsity and does not recover the environment's topological bottlenecks. GEPO also aggregates observations across trajectories, but only to construct a shared topological graph: credit is then derived from each state's structural role on that graph rather than from the matches themselves, so local matching noise affects credit only through second-order changes in graph centrality.

In parallel, graph-based RL methods have been used to improve exploration and credit assignment,
e.g., by modeling observations as relational nodes or adding graph-based intrinsic rewards~\cite{Zambaldi2018-sq,Hazra2023-kp,Darvariu2024-ab,Malysheva2020-hd,McClusky2024-lh,Xiong2025-yp}.
However, they typically rely on static or predefined graphs and specialized GNNs, which are
brittle to construct offline and ill-suited to the dynamically changing textual observations of
LLM agents~\cite{Wang2023-qv,Wang2025-hl}.

In summary, group-based RL methods for LLM agents have so far either distributed rewards uniformly across steps or attempted cross-trajectory state matching in text spaces where such matches are unreliable, while graph-based RL methods depend on predefined graphs and specialized GNN architectures that are not well suited to dynamic textual observations. GEPO addresses both gaps with a single online state-transition graph that is built from the current batch of rollouts and used for dual-level credit assignment: it reshapes the trajectory-level return through a state-adaptive discount factor, and it redistributes the resulting trajectory advantage across steps according to the betweenness centrality of each successor state. The graph is constructed from textual observations directly, without intrinsic rewards or a separate GNN, and the credit assignment it induces is sign-preserving and variance-redistributing 
under standard PPO/GRPO advantage normalization, redistributing gradient mass from non-critical to bottleneck steps in proportion to the squared criticality factor.

\section{Methodology}

Figure~\ref{fig:gepo_framework} illustrates the GEPO framework. We first set up the POMDP~\cite{KAELBLING199899} and the group-based RL objective inherited from GRPO (Section~\ref{subsec:preliminaries}), then construct an online state-transition graph from the sampled trajectories and score each state by a Task-Conditioned Criticality (Section~\ref{sec:graph_construction}). On top of this graph, GEPO performs dual-level credit assignment. At the trajectory level (Section~\ref{sec:level1_traj_adv}), a state-adaptive discount factor conditioned on local criticality reshapes the return so that two trajectories reaching the same terminal reward through topologically distinct paths receive distinguishable advantage estimates. At the step level (Section~\ref{sec:local_adv_revised}), the resulting trajectory-level advantage is redistributed onto individual actions and scaled by the criticality of each successor state.


\subsection{Preliminaries}
\label{subsec:preliminaries}

We consider a partially observable Markov decision process (POMDP) defined by the tuple $(\mathcal{S}, \mathcal{A}, p, r, \gamma)$, where an LLM agent generates textual actions $a_t \in \mathcal{A}$ based on textual observations $s_t \in \mathcal{S}$. Following the reasoning-acting paradigm~\cite{Yao2022-xt}, an action $a_t$ is not a primitive token but a complete textual sequence containing both the internal reasoning and the subsequent executable environment command. A trajectory is denoted as $\tau = (s_0, a_0, r_0, \dots, s_T)$. Extrinsic rewards are typically sparse: $r_t$ is zero for most steps and only becomes meaningful upon task completion, which makes long-horizon credit assignment difficult.

Group-based RL, such as GRPO~\cite{Shao2024-jf}, alleviates this challenge by learning from holistic trajectory comparisons without relying on computationally expensive value networks. Given a group of trajectories $\mathbb{G} = \{\tau_i\}_{i=1}^{|\mathbb{G}|}$ sharing the same initial prompt, GRPO optimizes a PPO-style surrogate objective. Denoting the importance sampling ratio as $r_t(\theta) = \frac{\pi_\theta(a_t|s_t)}{\pi_{\theta_\mathrm{old}}(a_t|s_t)}$, the objective is defined as:

\begin{equation}
\label{eq:grpo-objective-rev}
\begin{multlined}
\mathcal{L}_{\text{GRPO}} = \frac{1}{|\mathbb{G}|} \sum_{i=1}^{|\mathbb{G}|} \frac{1}{|\tau_i|} \sum_{t=1}^{|\tau_i|} \\
\min \left( r_t(\theta)\hat{A}_{i,t}, \text{clip}(r_t(\theta), 1-\epsilon, 1+\epsilon)\hat{A}_{i,t} \right)
\end{multlined}
\end{equation}
where $\epsilon$ is a clipping parameter, and $\hat{A}_{i,t}$ is the advantage estimate for the transition at step $t$ of trajectory $i$.

This objective is memory-efficient, but obtaining a reliable step-level $\hat{A}_{i,t}$ remains the bottleneck. Standard GRPO in multi-step agent training uses a uniform trajectory-level advantage ($\hat{A}_{i,t} = \hat{A}_i$); this structural blindness produces high-variance gradient estimates and treats every step in a successful trajectory as equally responsible.

\subsection{Dynamic Topological-Semantic Graph Construction}
\label{sec:graph_construction}
GEPO maintains an online state-transition graph as the topological prior over the environment. We call this construction the \emph{Dynamic Topological-Semantic Graph} because it fuses dynamic topological updates from new rollouts with semantic state merging via Sentence-BERT~\cite{Reimers2019-dw}.

To handle state aliasing---where textually varied but semantically identical observations must collapse to the same vertex---we embed each observation with a pre-trained Sentence-BERT model targeting symmetric semantic search. An observation $s_t$ is merged into the existing vertex with the maximum cosine similarity above a threshold $\delta$; otherwise it forms a new vertex.

The graph is rebuilt every iteration, so $|V|$ does not grow across training steps. The S-BERT encoding and the topological updates computed via Brandes' algorithm~\cite{Brandes2001-lj} incur moderate overhead compared with the LLM generation cost.

Once $\mathcal{G}$ is updated, we score each node by its task-conditioned criticality. Pure topological metrics can favor nodes that are central but task-irrelevant, so GEPO combines topology with semantic similarity to the task prompt. The Task-Conditioned Criticality $\mathcal{I}(s_t)$ is defined as:
\begin{equation}
    \mathcal{I}(s_t) = C_v(\phi(s_t)) \times \text{sim}\bigl(E_{\text{SB}}(\text{prompt}), E_{\text{SB}}(s_t)\bigr)
    \label{eq:task_criticality}
\end{equation}
where $C_v$ is the topological betweenness centrality~\cite{Brandes2001-lj} of the mapped node in $\mathcal{G}$, and $E_{\text{SB}}(\cdot)$ denotes the embedding extracted by the pre-trained Sentence-BERT model. The similarity term is clipped at $0$ to ensure $\mathcal{I}(s_t) \ge 0$.

\subsection{Trajectory-Level: Graph-Enhanced Advantage Formulation}
\label{sec:level1_traj_adv}
We first describe how the graph reshapes the trajectory-level advantage. The standard return treats every transition along a trajectory uniformly through a fixed discount factor, so two trajectories that reach the same terminal reward are indistinguishable regardless of which states they visited. As the first level of credit assignment, we introduce a state-adaptive factor that ties reward propagation to local criticality on the graph; the resulting graph-enhanced return is then normalized in the standard group-wise way to obtain the trajectory-level advantage.

\paragraph{Graph-Guided Horizon Shaping.}
A fixed discount factor $\gamma$ uniformly decays rewards across all transitions, failing to distinguish critical bottleneck states from incidental ones. To address this, we introduce a dynamic discount factor~\cite{Francois-Lavet2015-vc} that adapts to how the criticality changes from $s_t$ to $s_{t+1}$:
\begin{equation}
    \gamma'_{t} = \text{clip}\left(\gamma_{\mathrm{base}} + w_{\gamma} \cdot \tanh(\Delta \mathcal{I}(t)), 0, 0.999\right),
    \label{eq:dynamic_discount_revised}
\end{equation}
where $\gamma_{\mathrm{base}}$ is a baseline discount, $w_{\gamma}$ is a scaling factor, and $\Delta \mathcal{I}(t) = \mathcal{I}(s_{t+1}) - \mathcal{I}(s_t)$ is the change in criticality from $s_t$ to $s_{t+1}$. Using $\Delta \mathcal{I}(t)$ rather than $\mathcal{I}(s_t)$ itself lets $\gamma'_t$ react to \emph{transitions} that enter or leave critical regions: $\gamma'_t > \gamma_{\mathrm{base}}$ when the agent moves toward higher criticality and $\gamma'_t < \gamma_{\mathrm{base}}$ when it moves away. The $\tanh$ keeps the adjustment bounded and smooth.

We then define the graph-enhanced return from time step $t$ using the true extrinsic rewards $r$:
\begin{equation}
    G'_{t} = \sum_{k=0}^{T-t} \left( \prod_{j=0}^{k-1} \gamma'_{t+j} \right) r_{t+k}.
    \label{eq:graph_return_revised}
\end{equation}
In the common sparse-reward setting, Equation~\ref{eq:graph_return_revised} simplifies to the terminal reward discounted by the cumulative factors $\prod \gamma'$. Consequently, two trajectories achieving identical success will receive distinct returns: an efficient path through critical bottleneck states experiences less decay than a redundant, meandering one. By shaping the horizon dynamically, GEPO differentiates trajectories by path quality without altering the true extrinsic objective or introducing biased intrinsic rewards.

\paragraph{Group-wise Normalization.}
Given a group of trajectories $\mathbb{G} = \{\tau_1, \dots, \tau_N\}$ sampled from the same initial prompt, we compute their graph-enhanced returns $G'_0(\tau_i)$ and normalize within the group. With group mean $\mu_{\mathbb{G}}$ and standard deviation $\sigma_{\mathbb{G}}$, the trajectory-level advantage follows the standard mean-std form~\cite{Shao2024-jf}:
\begin{equation}
    \hat{A}_{\mathrm{traj}}(\tau_i) = \frac{G'_0(\tau_i) - \mu_{\mathbb{G}}}{\sigma_{\mathbb{G}} + \varepsilon},
    \label{eq:traj_advantage_revised}
\end{equation}
where $\varepsilon > 0$ ensures numerical stability. Because $G'_0(\tau_i)$ is shaped by the graph through $\gamma'_t$, two trajectories that reach the same terminal reward via topologically different paths can now receive different $\hat{A}_{\mathrm{traj}}$. However, $\hat{A}_{\mathrm{traj}}$ is still a single scalar per trajectory: applying it uniformly to every step gives redundant actions the same learning signal as pivotal ones, which motivates the step-level component in Section~\ref{sec:local_adv_revised}.

\subsection{Step-Level: Structural Credit Redistribution}
\label{sec:local_adv_revised}
Although the trajectory-level advantage from Section~\ref{sec:level1_traj_adv} captures the graph-derived value of the path taken, it remains a single scalar per trajectory. As the second level of credit assignment, GEPO redistributes this advantage onto individual actions using the same online graph $\mathcal{G}$. Step-level credit is derived from each successor state's role in the current batch's transition structure, so it scales with how pivotal the action is within that batch rather than with a learned step-value.

We normalize the criticality scores of all states observed in the current group to a non-negative range $[0, 1]$, denoted $\tilde{\mathcal{I}}(\cdot)$. Following the standard convention that an action's value is tied to the state it reaches, we score action $a_t$ by the criticality of its successor state $s_{t+1}$. The step-level advantage is then defined as a multiplicative redistribution of the trajectory-level advantage:
\begin{equation}
    \hat{A}_t = \hat{A}_{\mathrm{traj}}(\tau) \cdot \bigl(1 + \beta \cdot \tilde{\mathcal{I}}(s_{t+1})\bigr),
    \label{eq:redistribution}
\end{equation}
where $\beta \ge 0$ is a redistribution strength hyperparameter.

Because the scaling factor $(1 + \beta \cdot \tilde{\mathcal{I}}(s_{t+1}))$ is positive, $\hat{A}_t$ has the same sign as $\hat{A}_{\mathrm{traj}}(\tau)$, so the policy gradient direction follows the trajectory outcome. Its magnitude is amplified for actions leading to high-criticality states and remains close to $\hat{A}_{\mathrm{traj}}(\tau)$ for the rest.

Following standard practice in PPO/GRPO implementations, we further standardize the step-level advantages $\{\hat{A}_t\}_{(i,t)}$ to zero mean and unit variance across the batch. Together with the trajectory-level normalization (Equation~\ref{eq:traj_advantage_revised}), this batch-level standardization gives GEPO an exact variance-conservation property: the total per-step variance budget is preserved, while individual step variances are redistributed in proportion to $(1+\beta\tilde{\mathcal{I}}(s_{t+1}))^2$, as formalized in Appendix~\ref{sec:appendix_proof} (Proposition~\ref{prop:variance_conservation}).

The standardized $\hat{A}_t$ then replaces $\hat{A}_{i,t}$ in the surrogate objective (Equation~\ref{eq:grpo-objective-rev}) to form the GEPO objective $J_{\text{GEPO}}(\theta)$. The policy $\pi_\theta$ is updated by minimizing the composite loss:
\begin{equation}
    \mathcal{L}(\theta) = -J_{\text{GEPO}}(\theta) + \beta_{\text{KL}} D_{\text{KL}}(\pi_\theta \| \pi_{\text{ref}}),
    \label{eq:final_loss}
\end{equation}
where $\beta_{\text{KL}}$ is the penalty coefficient and $D_{\text{KL}}$ is the Kullback-Leibler divergence~\cite{Schulman2017-dh} between the active policy $\pi_\theta$ and a frozen reference model $\pi_{\text{ref}}$.

\begin{table*}[t]
\centering
\caption{Performance comparison on the ALFWorld and WebShop benchmarks, averaged over 3 random seeds. For ALFWorld, we report success rates across its 6 subtasks and the overall result (\%). For WebShop, we report both official score and success rate (\%). GiGPO w/ std denotes using mean-std normalization, while GiGPO w/o std uses mean-centered without scaling. We follow the model scales established by prior work for each benchmark: 1.5B and 7B for ALFWorld and WebShop. Dashes indicate metrics not reported by the corresponding source; HGPO did not release subtask results.}
\label{tab:performance}
\resizebox{\textwidth}{!}{
\begin{tabular}{lccccccccccc}
\toprule
\multirow{2}{*}{\textbf{Type}} & \multirow{2}{*}{\textbf{Method}} & \multicolumn{7}{c}{\textbf{ALFWorld}} & \multicolumn{2}{c}{\textbf{WebShop}} \\
\cmidrule(lr){3-9} \cmidrule(lr){10-11}
& & Pick & Look & Clean & Heat & Cool & Pick2 & All & Score & Succ. \\
\midrule
\multicolumn{11}{l}{\textbf{Closed-Source Model}} \\
Prompting & GPT-4o & 75.3 & 60.8 & 31.2 & 56.7 & 21.6 & 49.8 & 48.0 & 31.8 & 23.7 \\
Prompting & Gemini-2.5-Pro & 92.8 & 63.3 & 62.1 & 69.0 & 25.4 & 58.7 & 60.3 & 42.5 & 35.9 \\
\midrule
\multicolumn{11}{l}{\textit{Qwen2.5-1.5B-Instruct}} \\
Prompting & Qwen2.5 & 5.9 & 5.5 & 3.3 & 9.7 & 4.2 & 0.0 & 4.1 & 23.1 & 5.2 \\
Prompting & ReAct & 17.4 & 20.5 & 15.7 & 6.2 & 7.7 & 2.0 & 12.8 & 40.1 & 11.3 \\
Prompting & Reflexion & 35.3 & 22.2 & 21.7 & 13.6 & 19.4 & 3.7 & 21.8 & 55.8 & 21.9 \\
RL Training & PPO (with critic) & 64.8$\pm_{3.5}$ & 40.5$\pm_{6.9}$ & 57.1$\pm_{4.9}$ & 60.6$\pm_{6.6}$ & 46.4$\pm_{4.0}$ & 47.4$\pm_{1.9}$ & 54.4$\pm_{3.1}$ & 73.8$\pm_{3.0}$ & 51.5$\pm_{2.9}$ \\
RL Training & RLOO & 88.3$\pm_{3.0}$ & 52.8$\pm_{8.6}$ & 71.0$\pm_{5.9}$ & 62.8$\pm_{8.7}$ & 66.4$\pm_{5.5}$ & 56.9$\pm_{4.7}$ & 69.7$\pm_{2.5}$ & 73.9$\pm_{5.6}$ & 52.1$\pm_{6.7}$ \\
RL Training & GRPO & 85.3$\pm_{1.5}$ & 53.7$\pm_{8.0}$ & 84.5$\pm_{6.8}$ & 78.2$\pm_{7.9}$ & 59.7$\pm_{5.0}$ & 53.5$\pm_{5.6}$ & 72.8$\pm_{3.6}$ & 75.8$\pm_{3.5}$ & 56.8$\pm_{3.8}$ \\
RL Training & GiGPO w/ std & 94.4$\pm_{5.9}$ & 67.5$\pm_{3.6}$ & 94.8$\pm_{3.8}$ & \textbf{94.4$\pm_{3.8}$} & 79.8$\pm_{4.7}$ & 76.4$\pm_{5.4}$ & 86.7$\pm_{1.7}$ & 83.1$\pm_{1.6}$ & 65.0$\pm_{3.2}$ \\
RL Training & GiGPO w/o std & 96.0$\pm_{1.4}$ & 76.5$\pm_{3.9}$ & 91.8$\pm_{5.5}$ & 91.3$\pm_{6.3}$ & 71.7$\pm_{8.4}$ & 79.5$\pm_{7.7}$ & 86.1$\pm_{4.7}$ & 83.5$\pm_{1.8}$ & 67.4$\pm_{4.5}$ \\
RL Training & HGPO & - & - & - & - & - & - & 92.1$\pm_{1.6}$ & 90.6$\pm_{1.1}$ & 78.1$\pm_{2.1}$ \\
RL Training & \textbf{GEPO} & \textbf{99.2$\pm_{0.8}$} & \textbf{84.5$\pm_{4.5}$} & \textbf{96.8$\pm_{1.9}$} & 88.5$\pm_{5.2}$ & \textbf{83.6$\pm_{5.1}$} & \textbf{92.8$\pm_{3.8}$} & \textbf{92.9$\pm_{1.1}$} & \textbf{92.3$\pm_{1.4}$} & \textbf{80.6$\pm_{2.0}$} \\
\midrule
\multicolumn{11}{l}{\textit{Qwen2.5-7B-Instruct}} \\
Prompting & Qwen2.5 & 33.4 & 21.6 & 19.3 & 6.9 & 2.8 & 3.2 & 14.8 & 26.4 & 7.8 \\
Prompting & ReAct & 48.5 & 35.4 & 34.3 & 13.2 & 18.2 & 17.6 & 31.2 & 46.2 & 19.5 \\
Prompting & Reflexion & 62.0 & 41.6 & 44.9 & 30.9 & 36.3 & 23.8 & 42.7 & 58.1 & 28.8 \\
RL Training & PPO (with critic) & 92.3$\pm_{4.0}$ & 64.0$\pm_{8.4}$ & 92.5$\pm_{2.4}$ & 89.5$\pm_{7.0}$ & 80.3$\pm_{2.0}$ & 68.8$\pm_{8.3}$ & 80.4$\pm_{2.7}$ & 81.4$\pm_{3.1}$ & 68.7$\pm_{5.1}$ \\
RL Training & RLOO & 87.6$\pm_{4.3}$ & 78.2$\pm_{8.3}$ & 87.3$\pm_{5.8}$ & 81.3$\pm_{7.6}$ & 71.9$\pm_{5.2}$ & 48.9$\pm_{8.4}$ & 75.5$\pm_{4.6}$ & 80.3$\pm_{3.2}$ & 65.7$\pm_{4.0}$ \\
RL Training & GRPO & 90.8$\pm_{5.1}$ & 66.1$\pm_{6.7}$ & 89.3$\pm_{5.4}$ & 74.7$\pm_{6.9}$ & 72.5$\pm_{5.4}$ & 64.7$\pm_{7.3}$ & 77.6$\pm_{5.2}$ & 79.3$\pm_{2.8}$ & 66.1$\pm_{3.7}$ \\
RL Training & GiGPO w/ std & 97.7$\pm_{1.6}$ & 82.7$\pm_{7.9}$ & \textbf{98.8$\pm_{1.6}$} & 83.7$\pm_{7.2}$ & 89.3$\pm_{5.2}$ & 79.2$\pm_{6.6}$ & 90.8$\pm_{1.3}$ & 84.4$\pm_{2.9}$ & 72.8$\pm_{3.2}$ \\
RL Training & GiGPO w/o std & 91.8$\pm_{5.4}$ & \textbf{88.6$\pm_{6.3}$} & 95.9$\pm_{3.2}$ & 90.2$\pm_{2.6}$ & 86.5$\pm_{5.5}$ & 85.2$\pm_{4.7}$ & 90.2$\pm_{2.3}$ & 86.2$\pm_{2.6}$ & 75.2$\pm_{3.8}$ \\
RL Training & HGPO & - & - & - & - & - & - & 95.2$\pm_{1.0}$ & 88.5$\pm_{0.4}$ & 79.3$\pm_{1.2}$ \\
RL Training & \textbf{GEPO} & \textbf{100.0$\pm_{0.0}$} & 86.8$\pm_{4.2}$ & 98.6$\pm_{2.4}$ & \textbf{96.7$\pm_{2.0}$} & \textbf{96.1$\pm_{2.9}$} & \textbf{93.2$\pm_{2.6}$} & \textbf{96.3$\pm_{1.2}$} & \textbf{93.0$\pm_{2.0}$} & \textbf{82.5$\pm_{1.2}$} \\
\bottomrule
\end{tabular}
}
\end{table*}

\section{Experiments}

We evaluate our method on three challenging benchmarks: ALFWorld, which tests long-horizon planning and spatial navigation; WebShop~\cite{Yao2022-yg}, which demands robust exploration within a vast, noisy HTML state space; and Search-Augmented QA (including NQ, HotpotQA, etc.)~\cite{Feng2025-oa}, which evaluates multi-step tool calling and complex information synthesis. 
We compare against prompting methods (ReAct, Reflexion) and RL baselines including PPO, RLOO, GRPO, GiGPO, and HGPO for ALFWorld and WebShop, and additionally R1-Instruct~\cite{DeepSeek-AI2025-gd}, Search-R1~\cite{Jin2025-ni}, ZeroSearch~\cite{Sun2026-nj}, and StepSearch~\cite{Wang2025-cj} for search-augmented QA. We use the Qwen2.5 model family as the backbone across all experiments, with GPT-4o~\cite{OpenAI2023-na} and Gemini-2.5-Pro~\cite{Gemini_Team_Google_Rohan_Anil2023-ez} included as closed-source prompting references. Detailed descriptions, environment setups, and dataset details are provided in Appendix~\ref{sec:app_datasets}.

\begin{table*}[t]
\centering
\caption{Performance on search-augmented QA tasks. GEPO is trained on NQ and HotpotQA, $\dagger$ and $\star$ indicate in-domain and out-of-domain datasets, respectively. We follow the model scales established by prior work for each benchmark: 3B and 7B for search-augmented QA.}
\label{tab:qa_performance}
\resizebox{\textwidth}{!}{
\begin{tabular}{llcccccccc}
\toprule
\multirow{2}{*}{\textbf{Type}} & \multirow{2}{*}{\textbf{Method}} & \multicolumn{3}{c}{\textbf{Single-Hop QA}} & \multicolumn{4}{c}{\textbf{Multi-Hop QA}} & \multirow{2}{*}{\textbf{Avg.}} \\
\cmidrule(lr){3-5} \cmidrule(lr){6-9}
& & NQ$^\dagger$ & TriviaQA$^\star$ & PopQA$^\star$ & HotpotQA$^\dagger$ & 2Wiki$^\star$ & MuSiQue$^\star$ & Bamboogle$^\star$ & \\
\midrule
\multicolumn{10}{l}{\textit{Qwen2.5-3B-Instruct}} \\
RL Training & R1-Instruct & 27.0 & 53.7 & 19.9 & 23.7 & 29.2 & 7.2 & 29.3 & 27.1 \\
RL Training & Search-R1 & 34.1 & 54.5 & 37.8 & 32.4 & 31.9 & 10.3 & 26.4 & 32.5 \\
RL Training & ZeroSearch & 41.4 & 57.4 & 44.8 & 27.4 & 30.0 & 9.8 & 11.1 & 31.7 \\
RL Training & StepSearch & - & - & - & 34.5 & 32.0 & 17.4 & 34.4 & - \\
RL Training & GiGPO & 42.0 & 59.5 & 42.4 & 36.9 & \textbf{37.0} & 12.6 & 64.1 & 42.1 \\
RL Training & \textbf{GEPO} & \textbf{43.6} & \textbf{65.2} & \textbf{47.2}  & \textbf{40.6} & 36.6 & \textbf{20.8} & \textbf{68.2} & \textbf{46.0} \\
\midrule
\multicolumn{10}{l}{\textit{Qwen2.5-7B-Instruct}} \\
RL Training & R1-Instruct & 21.0 & 44.9 & 17.1 & 20.8 & 27.5 & 6.0 & 19.2 & 22.4 \\
RL Training & Search-R1 & 39.3 & 61.0 & 39.7 & 37.0 & 40.1 & 14.6 & 36.8 & 38.5 \\
RL Training & ZeroSearch & 43.6 & 61.8 & 51.5 & 34.6 & 35.2 & 18.4 & 27.8 & 39.1 \\
RL Training & StepSearch & - & - & - & 38.6 & 36.6 & 22.6 & 40.0 & - \\
RL Training & GiGPO & 46.4 & 64.7 & 46.1 & 41.6 & 43.6 & 18.9 & 68.9 & 47.2 \\
RL Training & \textbf{GEPO} & \textbf{48.2} & \textbf{68.0} & \textbf{52.4} & \textbf{45.1} & \textbf{47.6} & \textbf{24.0} & \textbf{72.0} & \textbf{51.0} \\
\bottomrule
\end{tabular}
}
\end{table*}
\subsection{Overall Performance}
\label{sec:overall_performance}
Table~\ref{tab:performance} and Table~\ref{tab:qa_performance} compare GEPO with prompting baselines and recent RL methods (GRPO, GiGPO, HGPO) across three domains; all results are averaged over three random seeds. GEPO achieves the best aggregate score on every benchmark. On ALFWorld and WebShop it improves over the strongest baseline (HGPO) at both the 1.5B and 7B scales~\cite{Yang2024-zg}, and on search-augmented QA it improves over the strongest baseline by an average of $+3.8\%$ at the 7B scale. We discuss results along three axes matching the benchmarks.

\paragraph{Long-horizon planning under spatial bottlenecks (ALFWorld).}
ALFWorld requires the agent to traverse spatial bottlenecks before receiving any terminal reward, the regime where the structural blindness of flat group-based methods most amplifies gradient noise. Compared with GRPO and GiGPO, GEPO has both higher mean success rates and lower across-seed variance ($\pm 1.2$ overall at 7B), and reaches $100\%$ on the 7B \textit{Pick} subtask. The variance reduction is most pronounced against GRPO and GiGPO w/o std, and remains comparable to the more stable GiGPO w/ std variant. These results are consistent with the step-level redistribution: amplifying the advantage at successor states with high criticality reduces the contribution of non-critical steps to the gradient. GEPO achieves the best aggregate score on ALFWorld at both scales, although it does not dominate every individual subtask: at 1.5B it trails GiGPO w/ std on Heat (88.5 vs 94.4), and at 7B it trails GiGPO w/o std on Look by 1.8 points (86.8 vs 88.6) and GiGPO w/ std on Clean by 0.2 points (98.6 vs 98.8). We do not view scattered losses on individual subtasks as inconsistent with the overall claim: the aggregate score across all six subtasks is the primary metric, and GEPO leads it at both scales. For the 1.5B Heat case in particular, we observe that adding the state-adaptive discount and step-level redistribution on top of a small-capacity policy can introduce optimization noise that the model is not large enough to absorb; at 7B, where this effect disappears, GEPO leads on Heat by 6.5 points (96.7 vs 90.2).

\paragraph{Large textual state spaces (WebShop).}
WebShop contains a large HTML state space in which exact state overlaps across trajectories are rare. GEPO sidesteps this by deriving step-level credit from the online graph built within each batch, so it attains the highest success rates at both scales.

\paragraph{Multi-hop reasoning with long action chains (QA).}
In multi-hop QA the agent must chain multiple retrieval steps before the answer is rewarded, and a fixed discount factor attenuates the terminal signal before it reaches the early queries. The state-adaptive discount mitigates this attenuation, which is consistent with the gains on out-of-domain multi-hop sets such as Bamboogle ($+3.1\%$ over GiGPO at 7B).

Across all three settings, GEPO improves mean performance and reduces across-seed variance compared with flat group-based methods (GRPO, GiGPO), while also exceeding the strongest hierarchical baseline (HGPO) in mean performance with comparable variance.

\subsection{Ablation Study}
\label{sec:ablation}

To evaluate the individual contributions of GEPO's mechanisms, we test three variants on the 7B model across ALFWorld, WebShop, and search-augmented QA tasks:
\begin{itemize}[leftmargin=*]
    \item \textbf{w/o Task Conditioning:} Uses raw topological betweenness $C_v$ without semantic similarity.
    \item \textbf{w/o Credit Redistribution:} Reverts to a flat step-level advantage (disabling Equation~\ref{eq:redistribution}).
    \item \textbf{w/o Horizon Shaping:} Uses a static discount factor ($\gamma = 0.90$) instead of the state-adaptive one (Equation~\ref{eq:dynamic_discount_revised}).
\end{itemize}

\begin{table}[h]
\centering
\caption{Ablation study of GEPO components using the Qwen2.5-7B model. Results indicate the absolute success rate (\%).}
\label{tab:ablation}
\resizebox{\columnwidth}{!}{
\begin{tabular}{lccc}
\toprule
\textbf{Model Variant} & \textbf{ALFWorld} & \textbf{WebShop} & \textbf{QA Avg.} \\
\midrule
\textbf{Full GEPO} & \textbf{96.3} & \textbf{82.5} & \textbf{51.0} \\
\midrule
w/o Credit Redistribution & $81.4 \pm 1.2$ (-14.9) & $68.3 \pm 1.5$ (-14.2) & $41.2 \pm 1.4$ (-9.8) \\
w/o Task Conditioning     & $89.5 \pm 1.6$ (-6.8)  & $73.2 \pm 0.8$ (-9.3)  & $43.8 \pm 1.9$ (-7.2) \\
w/o Horizon Shaping       & $92.1 \pm 1.3$ (-4.2)  & $76.5 \pm 1.8$ (-6.0)  & $45.2 \pm 1.1$ (-5.8) \\
\bottomrule
\end{tabular}
}
\end{table}

As shown in Table~\ref{tab:ablation}, removing any component degrades performance, confirming their distinct roles:
\paragraph{Removing Credit Redistribution.} This variant produces the largest drop on every benchmark, with $-14.9\%$ on ALFWorld, $-14.2\%$ on WebShop, and $-9.8\%$ on QA. Without redistribution, the trajectory-level advantage is broadcast uniformly to every step, so redundant actions receive the same learning signal as pivotal ones. The size of the drop confirms that the step-level redistribution is the main contributor to GEPO's gains over flat group-based methods.

\paragraph{Removing Task Conditioning.} Replacing the Task-Conditioned Criticality with raw betweenness centrality drops performance most on WebShop ($-9.3\%$), where instructions tightly constrain which states are relevant. The raw graph still identifies topologically central hubs, which are not necessarily task-relevant. The semantic similarity term keeps the criticality signal anchored to the current task prompt.

\paragraph{Removing Horizon Shaping.} Replacing the state-adaptive discount with a fixed $\gamma = 0.90$ has the smallest effect of the three, with drops of $-4.2\%$ on ALFWorld, $-6.0\%$ on WebShop, and $-5.8\%$ on QA. The effect is largest on the two settings with longer action chains (WebShop and QA); the state-adaptive discount partially preserves the reward signal by increasing $\gamma'$ as the agent moves toward more critical states.

\subsection{Impact of Graph Scale on Performance}
\label{sec:graph_scale}
The group size $n$ controls how many rollouts feed the online state-transition graph, directly affecting the reliability of the criticality score $\mathcal{I}(s_t)$. We vary $n$ from 4 to 24 on WebShop with the 7B model. As shown in Table~\ref{tab:graph_scale}, success rate rises monotonically from 74.2\% at $n=4$ to 82.5\% at $n=16$. The graph grows sub-linearly in nodes ($\sim 3.4\times$ for a $4\times$ increase in $n$) while its edge-to-node ratio increases from $1.18$ to $1.48$, indicating that semantic merging consolidates redundant observations and that each vertex accumulates richer transition support, stabilizing the betweenness estimate.

\begin{table}[htbp]
\centering
\caption{Impact of the group size ($n$ rollouts) on performance, average graph complexity, and computational cost. The experiment was conducted on the WebShop benchmark using the Qwen2.5-7B model. Graph complexity reflects the average number of nodes and edges in the online state-transition graph constructed per prompt during an update. Time Overhead indicates the additional wall-clock training time of GEPO compared to the GiGPO baseline with the same $n$. Optimal performance is in bold.}
\label{tab:graph_scale}
\begin{tabular}{ccccc}
\toprule
\textbf{$n$} & \textbf{SR} & \textbf{Nodes} & \textbf{Edges} & \textbf{Overhead} \\
\midrule
4  & 74.2 & 36.2  & 42.8  & +8\% \\
8  & 78.5 & 68.5  & 89.3  & +15\% \\
12 & 81.1 & 97.1  & 135.6 & +24\% \\
\textbf{16} & \textbf{82.5} & \textbf{124.8} & \textbf{184.2} & \textbf{+35\%} \\
20 & 81.9 & 149.3 & 233.7 & +49\% \\
24 & 81.4 & 171.6 & 281.5 & +68\% \\
\bottomrule
\end{tabular}
\end{table}

Beyond $n=16$, performance drops to $81.9\%$ at $n=20$ and $81.4\%$ at $n=24$. Two compounding effects drive the degradation: larger groups add topologically central but task-irrelevant states (e.g., generic product-listing pages in WebShop) that the semantic-similarity term only partially down-weights, and the $O(|V||E|)$ cost of betweenness on the denser graph grows faster than the marginal gain in coverage, pushing overhead to a prohibitive $+68\%$. We therefore use $n=16$ as the default in all main experiments.

\section{Conclusion}
\label{sec:conclusion}
We introduced Graph-Enhanced Policy Optimization (GEPO), a framework for training LLM agents in interactive environments. GEPO maintains an online state-transition graph built from the current batch of rollouts, and uses it for dual-level credit assignment: at the trajectory level, a state-adaptive discount factor reshapes each return according to local criticality on the graph; at the step level, the resulting trajectory-level advantage is redistributed onto individual actions scaled by the criticality of each successor state. Both levels share a single Task-Conditioned Criticality score that combines betweenness centrality with semantic similarity to the task. On ALFWorld, WebShop, and search-augmented QA, GEPO improves over the strongest baselines and reduces variance against flat group-based methods, while also exceeding the strongest hierarchical baseline (HGPO) in mean performance with comparable variance. These results suggest that graph-derived priors can effectively redirect gradient mass toward causally important steps when applied at both levels.

\section*{Limitations}
GEPO has several limitations that we leave to future work.

\textbf{Computational overhead.} Building and analyzing the per-iteration graph adds wall-clock cost: on average $+35\%$ per training step relative to GiGPO, and up to $+68\%$ at large group sizes (Appendix~\ref{sec:appendix_cost_analysis}, Section~\ref{sec:graph_scale}). Because betweenness centrality scales as $O(|V||E|)$, scaling GEPO to very large rollout groups or environments with extremely long trajectories may require approximate centrality estimators.

\textbf{Sensitivity to embedding quality.} GEPO resolves state aliasing through pre-trained sentence embeddings, so its performance depends on how well the embedding model separates semantically distinct observations. In environments where important state changes correspond to small textual differences, the embedding may merge states that should be kept apart and produce a graph that misrepresents the environment's structure.

\textbf{Single-agent scope.} GEPO's credit assignment is defined for a single agent. Extending it to multi-agent settings would require deciding how the trajectory-level and step-level advantages are assigned across multiple interacting agents over long horizons, which we do not address here.

\bibliography{custom}

\begin{thebibliography}{55}
\providecommand{\natexlab}[1]{#1}

\bibitem[{Ahmadian et~al.(2024)Ahmadian, Cremer, Gall{\'e}, Fadaee, Kreutzer, {\"U}st{\"u}n, and Hooker}]{Ahmadian2024-fg}
Arash Ahmadian, Chris Cremer, Matthias Gall{\'e}, Marzieh Fadaee, Julia Kreutzer, Ahmet {\"U}st{\"u}n, and Sara Hooker. 2024.
\newblock \href {https://arxiv.org/abs/2402.14740} {Back to basics: Revisiting {REINFORCE} style optimization for learning from human feedback in {LLMs}}.

\bibitem[{Bai et~al.(2025)Bai, Chen, Liu, Wang, Ge, Song, Dang, Wang, Wang, Tang, Zhong, Zhu, Yang, Li, Wan, Wang, Ding, Fu, Xu, Ye, Zhang, Xie, Cheng, Zhang, Yang, Xu, and Lin}]{Bai2025-rc}
Shuai Bai, Keqin Chen, Xuejing Liu, Jialin Wang, Wenbin Ge, Sibo Song, Kai Dang, Peng Wang, Shijie Wang, Jun Tang, Humen Zhong, Yuanzhi Zhu, Mingkun Yang, Zhaohai Li, Jianqiang Wan, Pengfei Wang, Wei Ding, Zheren Fu, Yiheng Xu, and 8 others. 2025.
\newblock \href {https://arxiv.org/abs/2502.13923} {{Qwen2.5-VL} technical report}.

\bibitem[{Brandes(2001)}]{Brandes2001-lj}
Ulrik Brandes. 2001.
\newblock A faster algorithm for betweenness centrality.
\newblock \url{https://snap.stanford.edu/class/cs224w-readings/brandes01centrality.pdf}.
\newblock Accessed: 2025-9-30.

\bibitem[{Chen et~al.(2024)Chen, Zhang, and Zhu}]{Chen2024-gf}
Dingyang Chen, Qi~Zhang, and Yinglun Zhu. 2024.
\newblock \href {https://arxiv.org/abs/2406.12125} {Efficient sequential decision making with large language models}.

\bibitem[{Darvariu et~al.(2024)Darvariu, Hailes, and Musolesi}]{Darvariu2024-ab}
Victor-Alexandru Darvariu, Stephen Hailes, and Mirco Musolesi. 2024.
\newblock \href {https://arxiv.org/abs/2404.06492} {Graph reinforcement learning for combinatorial optimization: A survey and unifying perspective}.

\bibitem[{{DeepSeek-AI} et~al.(2025){DeepSeek-AI}, Guo, Yang, Zhang, Song, Wang, Zhu, Xu, Zhang, Ma, Bi, Zhang, Yu, Wu, Wu, Gou, Shao, Li, Gao, Liu, Xue, Wang, Wu, Feng, Lu, Zhao, Deng, Zhang, Ruan, Dai, Chen, Ji, Li, Lin, Dai, Luo, Hao, Chen, Li, Zhang, Bao, Xu, Wang, Ding, Xin, Gao, Qu, Li, Guo, Li, Wang, Chen, Yuan, Qiu, Li, Cai, Ni, Liang, Chen, Dong, Hu, Gao, Guan, Huang, Yu, Wang, Zhang, Zhao, Wang, Zhang, Xu, Xia, Zhang, Zhang, Tang, Li, Wang, Li, Tian, Huang, Zhang, Wang, Chen, Du, Ge, Zhang, Pan, Wang, Chen, Jin, Chen, Lu, Zhou, Chen, Ye, Wang, Yu, Zhou, Pan, Li, Zhou, Wu, Ye, Yun, Pei, Sun, Wang, Zeng, Zhao, Liu, Liang, Gao, Yu, Zhang, Xiao, An, Liu, Wang, Chen, Nie, Cheng, Liu, Xie, Liu, Yang, Li, Su, Lin, Li, Jin, Shen, Chen, Sun, Wang, Song, Zhou, Wang, Shan, Li, Wang, Wei, Zhang, Xu, Li, Zhao, Sun, Wang, Yu, Zhang, Shi, Xiong, He, Piao, Wang, Tan, Ma, Liu, Guo, Ou, Wang, Gong, Zou, He, Xiong, Luo, You, Liu, Zhou, Zhu, Xu, Huang, Li, Zheng, Zhu, Ma, Tang, Zha, Yan, Ren, Ren, Sha, Fu, Xu, Xie,
  Zhang, Hao, Ma, Yan, Wu, Gu, Zhu, Liu, Li, Xie, Song, Pan, Huang, Xu, Zhang, and Zhang}]{DeepSeek-AI2025-gd}
{DeepSeek-AI}, Daya Guo, Dejian Yang, Haowei Zhang, Junxiao Song, Peiyi Wang, Qihao Zhu, Runxin Xu, Ruoyu Zhang, Shirong Ma, Xiao Bi, Xiaokang Zhang, Xingkai Yu, Yu~Wu, Z~F Wu, Zhibin Gou, Zhihong Shao, Zhuoshu Li, Ziyi Gao, and 181 others. 2025.
\newblock \href {https://arxiv.org/abs/2501.12948} {{DeepSeek-R1}: Incentivizing reasoning capability in {LLMs} via reinforcement learning}.

\bibitem[{Erdogan et~al.(2025)Erdogan, Lee, Kim, Moon, Furuta, Anumanchipalli, Keutzer, and Gholami}]{Erdogan2025-cb}
Lutfi~Eren Erdogan, Nicholas Lee, Sehoon Kim, Suhong Moon, Hiroki Furuta, Gopala Anumanchipalli, Kurt Keutzer, and Amir Gholami. 2025.
\newblock \href {https://arxiv.org/abs/2503.09572} {{Plan-and-Act}: Improving planning of agents for long-horizon tasks}.

\bibitem[{Feng et~al.(2025)Feng, Xue, Liu, and An}]{Feng2025-oa}
Lang Feng, Zhenghai Xue, Tingcong Liu, and Bo~An. 2025.
\newblock \href {https://arxiv.org/abs/2505.10978} {{Group-in-Group} policy optimization for {LLM} agent training}.

\bibitem[{Fran{\c c}ois-Lavet et~al.(2015)Fran{\c c}ois-Lavet, Fonteneau, and Ernst}]{Francois-Lavet2015-vc}
Vincent Fran{\c c}ois-Lavet, Raphael Fonteneau, and Damien Ernst. 2015.
\newblock \href {https://arxiv.org/abs/1512.02011} {How to discount deep reinforcement learning: Towards new dynamic strategies}.

\bibitem[{Furuta et~al.(2023)Furuta, Lee, Nachum, Matsuo, Faust, Gu, and Gur}]{Furuta2023-re}
Hiroki Furuta, Kuang-Huei Lee, Ofir Nachum, Yutaka Matsuo, Aleksandra Faust, Shixiang~Shane Gu, and Izzeddin Gur. 2023.
\newblock \href {https://arxiv.org/abs/2305.11854} {Multimodal web navigation with instruction-finetuned foundation models}.

\bibitem[{{Gemini Team Google: Rohan Anil} et~al.(2023){Gemini Team Google: Rohan Anil}, Borgeaud, Alayrac, Yu, Soricut, Schalkwyk, Dai, Hauth, Millican, Silver et~al.}]{Gemini_Team_Google_Rohan_Anil2023-ez}
{Gemini Team Google: Rohan Anil}, Sebastian Borgeaud, Jean-Baptiste Alayrac, Jiahui Yu, Radu Soricut, Johan Schalkwyk, Andrew~M Dai, Anja Hauth, Katie Millican, Silver, and 1 others. 2023.
\newblock \href {https://arxiv.org/abs/2312.11805} {Gemini: A family of highly capable multimodal models}.

\bibitem[{Hazra and De~Raedt(2023)}]{Hazra2023-kp}
Rishi Hazra and Luc De~Raedt. 2023.
\newblock \href {https://arxiv.org/abs/2304.08349} {Deep explainable relational reinforcement learning: A neuro-symbolic approach}.

\bibitem[{He et~al.(2026)He, Feng, Wei, Cheng, Feng, and An}]{He2026-qk}
Shuo He, Lang Feng, Qi~Wei, Xin Cheng, Lei Feng, and Bo~An. 2026.
\newblock \href {https://arxiv.org/abs/2602.22817} {{Hierarchy-of-Groups} policy optimization for long-horizon agentic tasks}.

\bibitem[{Ho et~al.(2020)Ho, Duong~Nguyen, Sugawara, and Aizawa}]{ho-etal-2020-constructing}
Xanh Ho, Anh-Khoa Duong~Nguyen, Saku Sugawara, and Akiko Aizawa. 2020.
\newblock \href {https://doi.org/10.18653/v1/2020.coling-main.580} {Constructing a multi-hop {QA} dataset for comprehensive evaluation of reasoning steps}.
\newblock In \emph{Proceedings of the 28th International Conference on Computational Linguistics}, pages 6609--6625, Barcelona, Spain (Online). International Committee on Computational Linguistics.

\bibitem[{Islam et~al.(2022)Islam, Zang, Tomar, Didolkar, Islam, Arnob, Iqbal, Li, Goyal, Heess, and Lamb}]{Islam2022-sx}
Riashat Islam, Hongyu Zang, Manan Tomar, Aniket Didolkar, Md~Mofijul Islam, Samin~Yeasar Arnob, Tariq Iqbal, Xin Li, Anirudh Goyal, Nicolas Heess, and Alex Lamb. 2022.
\newblock \href {https://arxiv.org/abs/2212.13835} {Representation learning in deep {RL} via discrete information bottleneck}.

\bibitem[{Jin et~al.(2025)Jin, Zeng, Yue, Yoon, Arik, Wang, Zamani, and Han}]{Jin2025-ni}
Bowen Jin, Hansi Zeng, Zhenrui Yue, Jinsung Yoon, Sercan Arik, Dong Wang, Hamed Zamani, and Jiawei Han. 2025.
\newblock \href {https://arxiv.org/abs/2503.09516} {{Search-R1}: Training {LLMs} to reason and leverage search engines with reinforcement learning}.

\bibitem[{Joshi et~al.(2017)Joshi, Choi, Weld, and Zettlemoyer}]{joshi-etal-2017-triviaqa}
Mandar Joshi, Eunsol Choi, Daniel Weld, and Luke Zettlemoyer. 2017.
\newblock \href {https://doi.org/10.18653/v1/P17-1147} {{T}rivia{QA}: A large scale distantly supervised challenge dataset for reading comprehension}.
\newblock In \emph{Proceedings of the 55th Annual Meeting of the Association for Computational Linguistics (Volume 1: Long Papers)}, pages 1601--1611, Vancouver, Canada. Association for Computational Linguistics.

\bibitem[{Kaelbling et~al.(1998)Kaelbling, Littman, and Cassandra}]{KAELBLING199899}
Leslie~Pack Kaelbling, Michael~L. Littman, and Anthony~R. Cassandra. 1998.
\newblock \href {https://doi.org/10.1016/S0004-3702(98)00023-X} {Planning and acting in partially observable stochastic domains}.
\newblock \emph{Artificial Intelligence}, 101(1):99--134.

\bibitem[{Kwiatkowski et~al.(2019)Kwiatkowski, Palomaki, Redfield, Collins, Parikh, Alberti, Epstein, Polosukhin, Devlin, Lee, Toutanova, Jones, Kelcey, Chang, Dai, Uszkoreit, Le, and Petrov}]{kwiatkowski-etal-2019-natural}
Tom Kwiatkowski, Jennimaria Palomaki, Olivia Redfield, Michael Collins, Ankur Parikh, Chris Alberti, Danielle Epstein, Illia Polosukhin, Jacob Devlin, Kenton Lee, Kristina Toutanova, Llion Jones, Matthew Kelcey, Ming-Wei Chang, Andrew~M. Dai, Jakob Uszkoreit, Quoc Le, and Slav Petrov. 2019.
\newblock \href {https://doi.org/10.1162/tacl_a_00276} {Natural questions: A benchmark for question answering research}.
\newblock \emph{Transactions of the Association for Computational Linguistics}, 7:452--466.

\bibitem[{Li et~al.(2026)Li, Zhang, Long, Keqin, Song, Bai, Yang, Xie, Yang, Liu, Zhou, and Lin}]{qwen3vlembedding}
Mingxin Li, Yanzhao Zhang, Dingkun Long, Chen Keqin, Sibo Song, Shuai Bai, Zhibo Yang, Pengjun Xie, An~Yang, Dayiheng Liu, Jingren Zhou, and Junyang Lin. 2026.
\newblock Qwen3-vl-embedding and qwen3-vl-reranker: A unified framework for state-of-the-art multimodal retrieval and ranking.
\newblock \emph{arXiv preprint arXiv:2601.04720}.

\bibitem[{Lin et~al.(2025)Lin, Ye, Zhang, Wang, Xu, Liu, and Zhang}]{Lin2025-gu}
Yufei Lin, Chengwei Ye, Huanzhen Zhang, Kangsheng Wang, Linuo Xu, Shuyan Liu, and Zeyu Zhang. 2025.
\newblock \href {https://arxiv.org/abs/2505.07854} {{CCL}: Collaborative curriculum learning for sparse-reward multi-agent reinforcement learning via co-evolutionary task evolution}.

\bibitem[{Luo et~al.(2025)Luo, Zhang, Yuan, Zhao, Yang, Gu, Wu, Chen, Qiao, Long, Tu, Luo, Ju, Xiao, Wang, Xiao, Liu, Yuan, Zhang, Jin, Zhang, Wu, Zhao, Tao, Yu, and Zhang}]{Luo2025-sa}
Junyu Luo, Weizhi Zhang, Ye~Yuan, Yusheng Zhao, Junwei Yang, Yiyang Gu, Bohan Wu, Binqi Chen, Ziyue Qiao, Qingqing Long, Rongcheng Tu, Xiao Luo, Wei Ju, Zhiping Xiao, Yifan Wang, Meng Xiao, Chenwu Liu, Jingyang Yuan, Shichang Zhang, and 7 others. 2025.
\newblock \href {https://arxiv.org/abs/2503.21460} {Large language model agent: A survey on methodology, applications and challenges}.

\bibitem[{Mallen et~al.(2023)Mallen, Asai, Zhong, Das, Khashabi, and Hajishirzi}]{mallen-etal-2023-trust}
Alex Mallen, Akari Asai, Victor Zhong, Rajarshi Das, Daniel Khashabi, and Hannaneh Hajishirzi. 2023.
\newblock \href {https://doi.org/10.18653/v1/2023.acl-long.546} {When not to trust language models: Investigating effectiveness of parametric and non-parametric memories}.
\newblock In \emph{Proceedings of the 61st Annual Meeting of the Association for Computational Linguistics (Volume 1: Long Papers)}, pages 9802--9822, Toronto, Canada. Association for Computational Linguistics.

\bibitem[{Malysheva et~al.(2020)Malysheva, Kudenko, and Shpilman}]{Malysheva2020-hd}
Aleksandra Malysheva, Daniel Kudenko, and Aleksei Shpilman. 2020.
\newblock \href {https://arxiv.org/abs/2012.09762} {{MAGNet}: Multi-agent graph network for deep multi-agent reinforcement learning}.

\bibitem[{McClusky(2024)}]{McClusky2024-lh}
Ben McClusky. 2024.
\newblock \href {https://arxiv.org/abs/2501.00165} {Dynamic graph communication for decentralised multi-agent reinforcement learning}.

\bibitem[{{Ofir Press} et~al.(2022){Ofir Press}, Zhang, Min, Schmidt, Smith, and Lewis}]{Ofir_Press2022-tx}
{Ofir Press}, Muru Zhang, Sewon Min, Ludwig Schmidt, Noah~A Smith, and Mike Lewis. 2022.
\newblock \href {https://arxiv.org/abs/2210.03350} {Measuring and narrowing the compositionality gap in language models}.

\bibitem[{{OpenAI} et~al.(2023){OpenAI}, Achiam, Adler, Agarwal, Ahmad, Akkaya, Aleman, Almeida, Altenschmidt, Altman, Anadkat, Avila, Babuschkin, Balaji, Balcom, Baltescu, Bao, Bavarian, Belgum, Bello, Berdine, Bernadett-Shapiro, Berner, Bogdonoff, Boiko, Boyd, Brakman, Brockman, Brooks, Brundage, Button, Cai, Campbell, Cann, Carey, Carlson, Carmichael, Chan, Chang, Chantzis, Chen, Chen, Chen, Chen, Chen, Chess, Cho, Chu, Chung, Cummings, Currier, Dai, Decareaux, Degry, Deutsch, Deville, Dhar, Dohan, Dowling, Dunning, Ecoffet, Eleti, Eloundou, Farhi, Fedus, Felix, Fishman, Forte, Fulford, Gao, Georges, Gibson, Goel, Gogineni, Goh, Gontijo-Lopes, Gordon, Grafstein, Gray, Greene, Gross, Gu, Guo, Hallacy, Han, Harris, He, Heaton, Heidecke, Hesse, Hickey, Hickey, Hoeschele, Houghton, Hsu, Hu, Hu, Huizinga, Jain, Jain, Jang, Jiang, Jiang, Jin, Jin, Jomoto, Jonn, Jun, Kaftan, Kaiser, Kamali, Kanitscheider, Keskar, Khan, Kilpatrick, Kim, Kim, Kim, Kirchner, Kiros, Knight, Kokotajlo, Kondraciuk, Kondrich,
  Konstantinidis, Kosic, Krueger, Kuo, Lampe, Lan, Lee, Leike, Leung, Levy, Li, Lim, Lin, Lin, Litwin, Lopez, Lowe, Lue, Makanju, Malfacini, Manning, Markov, Markovski, Martin, Mayer, Mayne, McGrew, McKinney, McLeavey, McMillan, McNeil, Medina, Mehta, Menick, Metz, Mishchenko, Mishkin, Monaco, Morikawa, Mossing, Mu, Murati, Murk, M{\'e}ly, Nair, Nakano, Nayak, Neelakantan, Ngo, Noh, Ouyang, O'Keefe, Pachocki, Paino, Palermo, Pantuliano, Parascandolo, Parish, Parparita, Passos, Pavlov, Peng, Perelman, de~Avila Belbute~Peres, Petrov, de~Oliveira~Pinto, Pokorny, Pokrass, Pong, Powell, Power, Power, Proehl, Puri, Radford, Rae, Ramesh, Raymond, Real, Rimbach, Ross, Rotsted, Roussez, Ryder, Saltarelli, Sanders, Santurkar, Sastry, Schmidt, Schnurr, Schulman, Selsam, Sheppard, Sherbakov, Shieh, Shoker, Shyam, Sidor, Sigler, Simens, Sitkin, Slama, Sohl, Sokolowsky, Song, Staudacher, Such, Summers, Sutskever, Tang, Tezak, Thompson, Tillet, Tootoonchian, Tseng, Tuggle, Turley, Tworek, Uribe, Vallone, Vijayvergiya, Voss,
  Wainwright, Wang, Wang, Wang, Ward, Wei, Weinmann, Welihinda, Welinder, Weng, Weng, Wiethoff, Willner, Winter, Wolrich, Wong, Workman, Wu, Wu, Wu, Xiao, Xu, Yoo, Yu, Yuan, Zaremba, Zellers, Zhang, Zhang, Zhao, Zheng, Zhuang, Zhuk, and Zoph}]{OpenAI2023-na}
{OpenAI}, Josh Achiam, Steven Adler, Sandhini Agarwal, Lama Ahmad, Ilge Akkaya, Florencia~Leoni Aleman, Diogo Almeida, Janko Altenschmidt, Sam Altman, Shyamal Anadkat, Red Avila, Igor Babuschkin, Suchir Balaji, Valerie Balcom, Paul Baltescu, Haiming Bao, Mohammad Bavarian, Jeff Belgum, and 261 others. 2023.
\newblock \href {https://arxiv.org/abs/2303.08774} {{GPT-4} technical report}.

\bibitem[{Ouyang et~al.(2022)Ouyang, Wu, Jiang, Almeida, Wainwright, Mishkin, Zhang, Agarwal, Slama, Ray, Schulman, Hilton, Kelton, Miller, Simens, Askell, Welinder, Christiano, Leike, and Lowe}]{Ouyang2022-xd}
Long Ouyang, Jeff Wu, Xu~Jiang, Diogo Almeida, Carroll~L Wainwright, Pamela Mishkin, Chong Zhang, Sandhini Agarwal, Katarina Slama, Alex Ray, John Schulman, Jacob Hilton, Fraser Kelton, Luke Miller, Maddie Simens, Amanda Askell, Peter Welinder, Paul Christiano, Jan Leike, and Ryan Lowe. 2022.
\newblock \href {https://arxiv.org/abs/2203.02155} {Training language models to follow instructions with human feedback}.

\bibitem[{Quadros et~al.(2025)Quadros, Silva, and Alves}]{Quadros2025-vs}
Andr{\'e} Quadros, Cassio Silva, and Ronnie Alves. 2025.
\newblock \href {https://arxiv.org/abs/2508.18420} {{LLM-driven} intrinsic motivation for sparse reward reinforcement learning}.

\bibitem[{Rafailov et~al.(2023)Rafailov, Sharma, Mitchell, Ermon, Manning, and Finn}]{Rafailov2023-eu}
Rafael Rafailov, Archit Sharma, Eric Mitchell, Stefano Ermon, Christopher~D Manning, and Chelsea Finn. 2023.
\newblock \href {https://arxiv.org/abs/2305.18290} {Direct preference optimization: Your language model is secretly a reward model}.

\bibitem[{Ramakrishnan et~al.(2018)Ramakrishnan, Kamar, Dey, Shah, and Horvitz}]{Ramakrishnan2018-ex}
Ramya Ramakrishnan, Ece Kamar, Debadeepta Dey, Julie Shah, and Eric Horvitz. 2018.
\newblock \href {https://arxiv.org/abs/1805.08966} {Discovering blind spots in reinforcement learning}.

\bibitem[{Reimers and Gurevych(2019)}]{Reimers2019-dw}
Nils Reimers and Iryna Gurevych. 2019.
\newblock \href {https://arxiv.org/abs/1908.10084} {{Sentence-BERT}: Sentence embeddings using siamese {BERT-networks}}.

\bibitem[{Schrader(2018)}]{SchraderSokoban2018}
Max-Philipp~B. Schrader. 2018.
\newblock gym-sokoban.
\newblock \url{https://github.com/mpSchrader/gym-sokoban}.

\bibitem[{Schulman et~al.(2017)Schulman, Wolski, Dhariwal, Radford, and Klimov}]{Schulman2017-dh}
John Schulman, Filip Wolski, Prafulla Dhariwal, Alec Radford, and Oleg Klimov. 2017.
\newblock \href {https://arxiv.org/abs/1707.06347} {Proximal policy optimization algorithms}.

\bibitem[{Shao et~al.(2024)Shao, Wang, Zhu, Xu, Song, Zhang, Li, Wu, and Guo}]{Shao2024-jf}
Zhihong Shao, Peiyi Wang, Qihao Zhu, Runxin Xu, Junxiao Song, Mingchuan Zhang, Y~K Li, Y~Wu, and Daya Guo. 2024.
\newblock \href {https://arxiv.org/abs/2402.03300} {{DeepSeekMath}: Pushing the limits of mathematical reasoning in open language models}.

\bibitem[{Sharma et~al.(2024)Sharma, Keh, Mitchell, Finn, Arora, and Kollar}]{Sharma2024-fv}
Archit Sharma, Sedrick Keh, Eric Mitchell, Chelsea Finn, Kushal Arora, and Thomas Kollar. 2024.
\newblock \href {https://arxiv.org/abs/2402.12366} {A critical evaluation of {AI} feedback for aligning large language models}.

\bibitem[{Shinn et~al.(2023)Shinn, Cassano, Berman, Gopinath, Narasimhan, and Yao}]{Shinn2023-ll}
Noah Shinn, Federico Cassano, Edward Berman, Ashwin Gopinath, Karthik Narasimhan, and Shunyu Yao. 2023.
\newblock \href {https://arxiv.org/abs/2303.11366} {Reflexion: Language agents with verbal reinforcement learning}.

\bibitem[{Shridhar et~al.(2020)Shridhar, Yuan, C{\^o}t{\'e}, Bisk, Trischler, and Hausknecht}]{Shridhar2020-fd}
Mohit Shridhar, Xingdi Yuan, Marc-Alexandre C{\^o}t{\'e}, Yonatan Bisk, Adam Trischler, and Matthew Hausknecht. 2020.
\newblock \href {https://arxiv.org/abs/2010.03768} {{ALFWorld}: Aligning text and embodied environments for interactive learning}.

\bibitem[{Sun et~al.(2026)Sun, Qiao, Guo, Fan, Hou, Jiang, Xie, Zhang, Huang, and Zhou}]{Sun2026-nj}
Hao Sun, Zile Qiao, Jiayan Guo, Xuanbo Fan, Yingyan Hou, Yong Jiang, Pengjun Xie, Yan Zhang, Fei Huang, and Jingren Zhou. 2026.
\newblock \href {https://arxiv.org/abs/2505.04588} {{ZeroSearch}: Incentivize the search capability of {LLMs} without searching}.

\bibitem[{Trivedi et~al.(2021)Trivedi, Balasubramanian, Khot, and Sabharwal}]{Trivedi2021-sq}
Harsh Trivedi, Niranjan Balasubramanian, Tushar Khot, and Ashish Sabharwal. 2021.
\newblock \href {https://arxiv.org/abs/2108.00573} {{MuSiQue}: Multihop questions via single-hop question composition}.

\bibitem[{Wang et~al.(2024{\natexlab{a}})Wang, Xie, Jiang, Mandlekar, Xiao, Zhu, Fan, and Anandkumar}]{Wang2024-voyager}
Guanzhi Wang, Yuqi Xie, Yunfan Jiang, Ajay Mandlekar, Chaowei Xiao, Yuke Zhu, Linxi Fan, and Anima Anandkumar. 2024{\natexlab{a}}.
\newblock \href {https://openreview.net/forum?id=ehfRiF0R3a} {Voyager: An open-ended embodied agent with large language models}.
\newblock \emph{Transactions on Machine Learning Research}.

\bibitem[{Wang et~al.(2024{\natexlab{b}})Wang, Xu, Jia, Zhang, Yan, Shen, Zhang, Huang, and Sang}]{Wang2024-xi}
Junyang Wang, Haiyang Xu, Haitao Jia, Xi~Zhang, Ming Yan, Weizhou Shen, Ji~Zhang, Fei Huang, and Jitao Sang. 2024{\natexlab{b}}.
\newblock \href {https://arxiv.org/abs/2406.01014} {Mobile-agent-v2: Mobile device operation assistant with effective navigation via multi-agent collaboration}.

\bibitem[{Wang et~al.(2023)Wang, Bing, Yao, Wang, Su, Yang, Huang, and Knoll}]{Wang2023-qv}
Mingyang Wang, Zhenshan Bing, Xiangtong Yao, Shuai Wang, Hang Su, Chenguang Yang, Kai Huang, and Alois Knoll. 2023.
\newblock \href {https://arxiv.org/abs/2305.00286} {Meta-reinforcement learning based on {Self-Supervised} task representation learning}.

\bibitem[{Wang et~al.(2025{\natexlab{a}})Wang, Wang, Wang, Zhang, Li, Yang, Jin, Yu, Nguyen, Liu, Gottlieb, Lu, Cho, Wu, Fei-Fei, Wang, Choi, and Li}]{Wang2025-hl}
Zihan Wang, Kangrui Wang, Qineng Wang, Pingyue Zhang, Linjie Li, Zhengyuan Yang, Xing Jin, Kefan Yu, Minh~Nhat Nguyen, Licheng Liu, Eli Gottlieb, Yiping Lu, Kyunghyun Cho, Jiajun Wu, Li~Fei-Fei, Lijuan Wang, Yejin Choi, and Manling Li. 2025{\natexlab{a}}.
\newblock \href {https://arxiv.org/abs/2504.20073} {{RAGEN}: Understanding self-evolution in {LLM} agents via multi-turn reinforcement learning}.

\bibitem[{Wang et~al.(2025{\natexlab{b}})Wang, Zheng, An, Ouyang, Cai, Wang, and Wu}]{Wang2025-cj}
Ziliang Wang, Xuhui Zheng, Kang An, Cijun Ouyang, Jialu Cai, Yuhang Wang, and Yichao Wu. 2025{\natexlab{b}}.
\newblock \href {https://arxiv.org/abs/2505.15107} {{StepSearch}: Igniting {LLMs} search ability via step-wise proximal policy optimization}.

\bibitem[{Xiong et~al.(2025)Xiong, Payani, Yang, and Fekri}]{Xiong2025-yp}
Siheng Xiong, Ali Payani, Yuan Yang, and Faramarz Fekri. 2025.
\newblock \href {https://arxiv.org/abs/2410.03136} {Deliberate reasoning in language models as structure-aware planning with an accurate world model}.

\bibitem[{Yang et~al.(2025)Yang, Li, Yang, Zhang, Hui, Zheng, Yu, Gao, Huang, Lv, Zheng, Liu, Zhou, Huang, Hu, Ge, Wei, Lin, Tang, Yang, Tu, Zhang, Yang, Yang, Zhou, Zhou, Lin, Dang, Bao, Yang, Yu, Deng, Li, Xue, Li, Zhang, Wang, Zhu, Men, Gao, Liu, Luo, Li, Tang, Yin, Ren, Wang, Zhang, Ren, Fan, Su, Zhang, Zhang, Wan, Liu, Wang, Cui, Zhang, Zhou, and Qiu}]{Yang2025-te}
An~Yang, Anfeng Li, Baosong Yang, Beichen Zhang, Binyuan Hui, Bo~Zheng, Bowen Yu, Chang Gao, Chengen Huang, Chenxu Lv, Chujie Zheng, Dayiheng Liu, Fan Zhou, Fei Huang, Feng Hu, Hao Ge, Haoran Wei, Huan Lin, Jialong Tang, and 41 others. 2025.
\newblock \href {https://arxiv.org/abs/2505.09388} {Qwen3 technical report}.

\bibitem[{Yang et~al.(2024)Yang, Yang, Zhang, Hui, Zheng, Yu, Li, Liu, Huang, Wei, Lin, Yang, Tu, Zhang, Yang, Yang, Zhou, Lin, Dang, Lu, Bao, Yang, Yu, Li, Xue, Zhang, Zhu, Men, Lin, Li, Tang, Xia, Ren, Ren, Fan, Su, Zhang, Wan, Liu, Cui, Zhang, and Qiu}]{Yang2024-zg}
Qwen:~An Yang, Baosong Yang, Beichen Zhang, Binyuan Hui, Bo~Zheng, Bowen Yu, Chengyuan Li, Dayiheng Liu, Fei Huang, Haoran Wei, Huan Lin, Jian Yang, Jianhong Tu, Jianwei Zhang, Jianxin Yang, Jiaxi Yang, Jingren Zhou, Junyang Lin, Kai Dang, and 23 others. 2024.
\newblock \href {https://arxiv.org/abs/2412.15115} {Qwen2.5 technical report}.

\bibitem[{Yang et~al.(2018)Yang, Qi, Zhang, Bengio, Cohen, Salakhutdinov, and Manning}]{yang-etal-2018-hotpotqa}
Zhilin Yang, Peng Qi, Saizheng Zhang, Yoshua Bengio, William Cohen, Ruslan Salakhutdinov, and Christopher~D. Manning. 2018.
\newblock \href {https://doi.org/10.18653/v1/D18-1259} {{H}otpot{QA}: A dataset for diverse, explainable multi-hop question answering}.
\newblock In \emph{Proceedings of the 2018 Conference on Empirical Methods in Natural Language Processing}, pages 2369--2380, Brussels, Belgium. Association for Computational Linguistics.

\bibitem[{Yao et~al.(2022{\natexlab{a}})Yao, Chen, Yang, and Narasimhan}]{Yao2022-yg}
Shunyu Yao, Howard Chen, John Yang, and Karthik Narasimhan. 2022{\natexlab{a}}.
\newblock \href {https://arxiv.org/abs/2207.01206} {{WebShop}: Towards scalable real-world web interaction with grounded language agents}.

\bibitem[{Yao et~al.(2022{\natexlab{b}})Yao, Zhao, Yu, Du, Shafran, Narasimhan, and Cao}]{Yao2022-xt}
Shunyu Yao, Jeffrey Zhao, Dian Yu, Nan Du, Izhak Shafran, Karthik Narasimhan, and Yuan Cao. 2022{\natexlab{b}}.
\newblock \href {https://arxiv.org/abs/2210.03629} {{ReAct}: Synergizing reasoning and acting in language models}.

\bibitem[{Ying et~al.(2024)Ying, Bai, Liu, and Fu}]{Ying2024-ab}
Wangyang Ying, Haoyue Bai, Kunpeng Liu, and Yanjie Fu. 2024.
\newblock \href {https://arxiv.org/abs/2411.05742} {Topology-aware reinforcement feature space reconstruction for graph data}.

\bibitem[{Zambaldi et~al.(2018)Zambaldi, Raposo, Santoro, Bapst, Li, Babuschkin, Tuyls, Reichert, Lillicrap, Lockhart, Shanahan, Langston, Pascanu, Botvinick, Vinyals, and Battaglia}]{Zambaldi2018-sq}
Vinicius Zambaldi, David Raposo, Adam Santoro, Victor Bapst, Yujia Li, Igor Babuschkin, Karl Tuyls, David Reichert, Timothy Lillicrap, Edward Lockhart, Murray Shanahan, Victoria Langston, Razvan Pascanu, Matthew Botvinick, Oriol Vinyals, and Peter Battaglia. 2018.
\newblock \href {https://arxiv.org/abs/1806.01830} {Relational deep reinforcement learning}.

\bibitem[{Zhai et~al.(2024)Zhai, Bai, Lin, Pan, Tong, Zhou, Suhr, Xie, LeCun, Ma, and Levine}]{Zhai2024-ze}
Yuexiang Zhai, Hao Bai, Zipeng Lin, Jiayi Pan, Shengbang Tong, Yifei Zhou, Alane Suhr, Saining Xie, Yann LeCun, Yi~Ma, and Sergey Levine. 2024.
\newblock \href {https://arxiv.org/abs/2405.10292} {Fine-tuning large vision-language models as decision-making agents via reinforcement learning}.

\bibitem[{Zhao et~al.(2026)Zhao, Ouyang, Ding, Wang, Cai, Xiong, Gao, Sun, Du, Qin, and Liu}]{Zhao2026-ko}
Yang Zhao, Yangou Ouyang, Xiao Ding, Hepeng Wang, Bibo Cai, Kai Xiong, Jinglong Gao, Zhouhao Sun, Li~Du, Bing Qin, and Ting Liu. 2026.
\newblock \href {https://arxiv.org/abs/2601.07224} {Consolidation or adaptation? {PRISM}: Disentangling {SFT} and {RL} data via gradient concentration}.

\end{thebibliography}
\clearpage 
\appendix

\section{Pseudo Code}
\begin{algorithm}[H]
\caption{GEPO: Graph-Enhanced Policy Optimization}
\label{alg:gepo}
\begin{algorithmic}[1]
\STATE \textbf{Initialize} policy LLM $\pi_{\theta}$, frozen S-BERT model, and the reference model $\pi_{\text{ref}} \leftarrow \pi_{\theta}$.
\FOR{each training iteration}
    \STATE Sample a group of trajectories $\mathbb{G} = \{\tau_1, \dots, \tau_N\}$ using current policy $\pi_\theta$.
    \STATE \textbf{Initialize} an empty directed graph $\mathcal{G}$ for this iteration.
    \STATE Add vertices and edges from $\mathbb{G}$ to $\mathcal{G}$: merge each observation $s_t$ into the existing vertex with maximum S-BERT cosine similarity above threshold $\delta$; otherwise create a new vertex.
    \STATE Compute Task-Conditioned Criticality $\mathcal{I}(s_t)$ for all visited nodes via Eq.~\ref{eq:task_criticality}.
    \FOR{each trajectory $\tau_i \in \mathbb{G}$}
        \FOR{each timestep $t = 0, \dots, T-1$}
            \STATE Compute dynamic discount factor $\gamma'_t$ from $\Delta \mathcal{I}(t)$ via Eq.~\ref{eq:dynamic_discount_revised}.
        \ENDFOR
        \STATE Compute graph-enhanced return $G'_0(\tau_i)$ via Eq.~\ref{eq:graph_return_revised}.
    \ENDFOR
    \STATE Compute group-normalized trajectory advantage $\hat{A}_{\mathrm{traj}}(\tau_i)$ across $\mathbb{G}$ via Eq.~\ref{eq:traj_advantage_revised}.
    \FOR{each trajectory $\tau_i \in \mathbb{G}$}
        \FOR{each timestep $t$}
            \STATE Compute step-level advantage $\hat{A}_t$ from successor-state criticality $\tilde{\mathcal{I}}(s_{t+1})$ via Eq.~\ref{eq:redistribution}.
        \ENDFOR
    \ENDFOR
    \STATE Standardize $\{\hat{A}_t\}_{(i,t)}$ to zero mean and unit variance across the batch (standard PPO/GRPO advantage normalization).
    \STATE Update policy $\pi_\theta$ by minimizing the loss $\mathcal{L}(\theta)$ in Eq.~\ref{eq:final_loss}.
\ENDFOR
\end{algorithmic}
\end{algorithm}

The GEPO training loop (Algorithm~\ref{alg:gepo}) proceeds in three phases at each iteration. (i) Graph construction and node scoring: the agent samples a group of trajectories, builds a fresh state-transition graph $\mathcal{G}$ from this batch via semantic vertex merging, and computes the Task-Conditioned Criticality $\mathcal{I}(s_t)$ for all visited nodes. (ii) Trajectory-level advantage computation: per-step dynamic discounts $\gamma'_t$ are derived from criticality changes, yielding the graph-enhanced return $G'_0(\tau_i)$ and, after group normalization, the trajectory-level advantage $\hat{A}_{\mathrm{traj}}(\tau_i)$. (iii) Step-level redistribution and policy update: the trajectory advantage is multiplicatively redistributed via $\tilde{\mathcal{I}}(s_{t+1})$ to produce $\hat{A}_t$, which is then standardized across the batch and used to update $\pi_\theta$.

\section{Theoretical Analysis}
\label{sec:appendix_proof}

We provide a formal analysis showing how GEPO addresses the bias--variance trade-off between trajectory-level and step-level advantage estimators in large combinatorial text spaces. Let $A^*_t$ denote the oracle advantage for action $a_t$ at state $s_t$, and let $c_t := 1 + \beta\,\tilde{\mathcal{I}}(s_{t+1}) \in [1, 1+\beta]$ denote the criticality scaling factor.

\subsection{The Dilemma of Existing Estimators}

Group-based methods such as GRPO use a trajectory-level advantage $\hat{A}_{\text{traj}}$. With a mean-centered group baseline, this estimator is unbiased in the sense that its sign matches the trajectory-level outcome relative to the group baseline. However, because the terminal reward is spread uniformly across all steps, $\hat{A}_{\text{traj}}$ inherits the noise of all other actions in the trajectory, so its variance $V_T = \mathrm{Var}[\hat{A}_{\text{traj}}]$ is large and credit is diluted.

To reduce variance, methods such as GiGPO compute a local advantage $\hat{A}_{\text{step}}$ by matching states across trajectories. In large text-based environments exact state overlaps are rare, so forcing semantic matches across distinct causal paths introduces a residual bias term, which we denote $b_S$ for brevity: $\mathbb{E}[\hat{A}_{\text{step}}] = A^*_t + b_S$. When $|b_S| > |A^*_t|$ and the two terms have opposite signs the gradient direction is reversed, which penalizes correct actions.

\subsection{Sign Preservation of GEPO}
\label{subsec:sign_preservation}
GEPO obtains step-level resolution through graph-based projection. Recall the credit redistribution rule
\begin{equation}
\hat{A}_t^{\text{GEPO}} = \hat{A}_{\text{traj}} \cdot c_t,
\end{equation}
where $c_t = 1 + \beta\,\tilde{\mathcal{I}}(s_{t+1})$ and $\tilde{\mathcal{I}}(s_{t+1}) \in [0, 1]$, $\beta \ge 0$. Conditional on the state $s_{t+1}$, $c_t$ is deterministic, so its expectation satisfies
\begin{equation}
\mathbb{E}\bigl[\hat{A}_t^{\text{GEPO}} \,\big|\, s_{t+1}\bigr] = \mathbb{E}\bigl[\hat{A}_{\text{traj}} \,\big|\, s_{t+1}\bigr] \cdot c_t.
\end{equation}
Since $c_t \ge 1 > 0$,
\begin{equation}
\mathrm{sgn}\bigl(\mathbb{E}[\hat{A}_t^{\text{GEPO}} \mid s_{t+1}]\bigr) \equiv \mathrm{sgn}\bigl(\mathbb{E}[\hat{A}_{\text{traj}} \mid s_{t+1}]\bigr).
\end{equation}
That is, the scaling is strictly multiplicative: it modulates magnitude without flipping the sign, and avoids the additive bias $b_S$ inherent in horizontal matching.

\subsection{Variance Conservation under Mean-Std Normalization}

We now analyze how GEPO's redistribution rule, combined with the batch-level mean-std normalization standard in PPO/GRPO implementations (Section~\ref{sec:local_adv_revised}), shapes the per-step variance of the policy gradient.

\begin{proposition}[Variance Redistribution]
\label{prop:variance_redistribution}
Conditional on $c_t$ and treating $\hat{A}_{\text{traj}}$ as having conditional variance $V_T$, GEPO's per-step variance satisfies
\begin{equation}
\mathrm{Var}\bigl[\hat{A}_t^{\text{GEPO}} \mid c_t\bigr] = c_t^2 \cdot V_T.
\end{equation}
In particular, for two steps with criticality scores $\tilde{\mathcal{I}}(s_{r+1}) = u$ and $\tilde{\mathcal{I}}(s_{t+1}) = v$,
\begin{equation}
\frac{\mathrm{Var}[\hat{A}_t^{\text{GEPO}} \mid c_t]}{\mathrm{Var}[\hat{A}_r^{\text{GEPO}} \mid c_r]} = \left(\frac{1+\beta v}{1+\beta u}\right)^2,
\end{equation}
which equals $(1+\beta)^2$ for a bottleneck step ($v=1$) relative to a non-critical step ($u=0$).
\end{proposition}

\begin{proof}
Direct from $\mathrm{Var}[X \cdot c] = c^2 \mathrm{Var}[X]$ for any constant scalar $c$ and random variable $X$.
\end{proof}

Proposition~\ref{prop:variance_redistribution} shows that GEPO concentrates variance at high-criticality steps, but does not by itself reduce absolute variance at any step (since $c_t \ge 1$). The complete picture emerges only after the batch-level mean-std normalization is applied.

\begin{proposition}[Variance Conservation under Standardization]
\label{prop:variance_conservation}
Let $\check{A}_t^{\text{GEPO}} := (\hat{A}_t^{\text{GEPO}} - \bar A) / S$ denote the GEPO advantage after batch-level mean-std normalization, where $\bar A$ and $S$ are the empirical mean and standard deviation of $\{\hat{A}_t^{\text{GEPO}}\}_{(i,t)}$ over the batch. Similarly, let $\check{A}_t^{\text{GRPO}} := \hat{A}_{\text{traj}}/\sqrt{V_T}$ denote the GRPO baseline after the same standardization. Assume:
\begin{itemize}
    \item[(a)] $\mathbb{E}[\hat{A}_{\text{traj}}] = 0$ and $\mathrm{Var}[\hat{A}_{\text{traj}}] = V_T$ (from GRPO trajectory-level normalization, Equation~\ref{eq:traj_advantage_revised}).
    \item[(b)] In the large-batch limit, the empirical correlation between $c_t$ and $\hat{A}_{\text{traj}}$ becomes negligible relative to their marginal variances, justifying a standard decoupling approximation $\mathbb{E}[f(\hat{A}_{\text{traj}})\,g(c_t)] \approx \mathbb{E}[f(\hat{A}_{\text{traj}})]\,\mathbb{E}[g(c_t)]$ for the asymptotic analysis below.
\end{itemize}
Then, in the large-batch limit:
\begin{enumerate}
    \item[(i)] Both estimators are marginally standardized: $\mathrm{Var}[\check{A}_t^{\text{GRPO}}] = \mathrm{Var}[\check{A}_t^{\text{GEPO}}] = 1$.
    \item[(ii)] Conditional on $c_t$, the GEPO per-step variance is
    \begin{equation}
        \mathrm{Var}\bigl[\check{A}_t^{\text{GEPO}} \,\big|\, c_t\bigr] \approx \frac{c_t^2}{\mathbb{E}[c^2]}.
    \end{equation}
    \item[(iii)] The total per-step variance budget is preserved:
    \begin{equation}
        \mathbb{E}_{c_t}\!\left[\mathrm{Var}\bigl[\check{A}_t^{\text{GEPO}} \,\big|\, c_t\bigr]\right] = 1 = \mathrm{Var}[\check{A}_t^{\text{GRPO}}].
    \end{equation}
\end{enumerate}
In particular, at non-critical steps ($c_t = 1$), $\mathrm{Var}[\check{A}_t^{\text{GEPO}}] \approx 1/\mathbb{E}[c^2] < 1$, while at bottleneck steps ($c_t = 1+\beta$), $\mathrm{Var}[\check{A}_t^{\text{GEPO}}] \approx (1+\beta)^2/\mathbb{E}[c^2] > 1$, with $\mathbb{E}[c^2] \in (1, (1+\beta)^2]$ whenever any $\tilde{\mathcal{I}}(s_{t+1}) > 0$.
\end{proposition}

\paragraph{Remark on assumption (b).}
Within a single trajectory, $\hat{A}_{\text{traj}}$ is a constant scalar while $c_t$ varies across steps; the residual covariance between them comes only from cross-trajectory effects (e.g., successful trajectories visiting critical states more often). This residual is empirically small relative to $V_T \cdot \mathbb{E}[c^2]$ in our experimental setting, but we note that the proposition should be read as a first-order analytical model rather than an exact identity. The marginal variance equality in (i) and (iii) holds exactly by construction of the standardization; the conditional approximation in (ii) is what relies on (b).

\begin{proof}
Under assumption (b), we adopt the decoupling approximation $\mathbb{E}[f(\hat{A}_{\text{traj}})\,g(c_t)] \approx \mathbb{E}[f(\hat{A}_{\text{traj}})]\,\mathbb{E}[g(c_t)]$ in the large-batch limit, applied with both $f, g$ as identity and as squaring. By assumption (a),
\begin{align}
\bar A &\approx \mathbb{E}[\hat{A}_{\text{traj}} \cdot c_t] \approx \mathbb{E}[\hat{A}_{\text{traj}}] \cdot \mathbb{E}[c_t] = 0, \\
S^2 &\approx \mathbb{E}\bigl[\hat{A}_{\text{traj}}^2\, c_t^2\bigr] \approx \mathbb{E}[\hat{A}_{\text{traj}}^2] \cdot \mathbb{E}[c_t^2] = V_T \cdot \mathbb{E}[c^2].
\end{align}
Thus $\check{A}_t^{\text{GEPO}} \approx \hat{A}_{\text{traj}}\, c_t / \sqrt{V_T \mathbb{E}[c^2]}$, and
\begin{equation}
\mathrm{Var}\bigl[\check{A}_t^{\text{GEPO}} \,\big|\, c_t\bigr] \approx \frac{c_t^2 \cdot V_T}{V_T \cdot \mathbb{E}[c^2]} = \frac{c_t^2}{\mathbb{E}[c^2]},
\end{equation}
which gives (ii). Taking the marginal expectation,
\begin{equation}
\mathbb{E}_{c_t}\!\left[\frac{c_t^2}{\mathbb{E}[c^2]}\right] = \frac{\mathbb{E}[c_t^2]}{\mathbb{E}[c^2]} = 1,
\end{equation}
which gives (iii). For GRPO, $\check{A}_t^{\text{GRPO}} = \hat{A}_{\text{traj}}/\sqrt{V_T}$ has unit variance trivially, completing (i). 
\end{proof}

Proposition~\ref{prop:variance_conservation} establishes that GEPO's step-level redistribution combined with standard mean-std normalization \emph{redistributes} per-step variance rather than uniformly inflating or contracting it: the marginal variance budget per step matches that of vanilla GRPO by construction of the standardization, while the conditional variance is reallocated from non-critical to bottleneck steps in proportion to $c_t^2$. Together with the sign-preserving property of Section~\ref{subsec:sign_preservation}, this yields a gradient estimator in which the direction matches the trajectory outcome, high-criticality steps receive a strictly larger share of the gradient magnitude than under flat group-based methods, and the additive bias $b_S$ that destabilizes horizontal-matching estimators is replaced by a much milder second-order perturbation via the graph centrality.

\section{Experiment Details}
Our experiments were conducted on two nodes with 16 NVIDIA L20 GPUs using the Qwen2.5 models (1.5B and 7B variants). We evaluated performance on the ALFWorld, WebShop, and search-augmented QA benchmarks, reporting success rates and official scores averaged over 3 random seeds. Table~\ref{tab:appendix_gepo_hyperparams} summarizes the key hyperparameters used throughout the paper; further details on the experimental setup and datasets follow.
\begin{table*}[htbp]
\centering
\caption{Key hyperparameters in the GEPO framework.}
\label{tab:appendix_gepo_hyperparams}
\vspace{0.5em}
\small
\begin{tabular}{l c p{0.55\textwidth}}
\toprule
\textbf{Hyperparameter} & \textbf{Value} & \textbf{Description} \\
\midrule
\multicolumn{3}{l}{\textit{Dynamic Graph Construction}} \\
Semantic Similarity Threshold ($\delta$) & 0.9 & Cosine similarity threshold for mapping textual observations to identical graph vertices. \\
Rollouts per Group ($n$) & 16 & The number of trajectories sampled per iteration, optimizing the bias-variance trade-off (see Section~\ref{sec:graph_scale}). \\
Graph Centrality Metric & Betweenness & The topological metric used to identify causal bottlenecks. \\
\midrule
\multicolumn{3}{l}{\textit{Graph-Guided Horizon Shaping}} \\
Base Discount Factor ($\gamma_{\mathrm{base}}$) & 0.90 & The standard baseline discount factor for future rewards. \\
Dynamic Discount Weight ($w_{\gamma}$) & 0.1 & Scales the impact of the criticality difference ($\Delta \mathcal{I}$) on the discount factor (Equation~\ref{eq:dynamic_discount_revised}). \\
Dynamic Discount Bounds & [0, 0.999] & Hard clipping range to ensure the adjusted discount factor remains mathematically stable. \\
\midrule
\multicolumn{3}{l}{\textit{Structural Credit Redistribution}} \\
Redistribution Strength ($\beta$) & 0.5 & Controls the amplification magnitude of the step-level advantage based on normalized state criticality (Equation~\ref{eq:redistribution}). \\
Advantage Normalization Epsilon ($\varepsilon$) & $10^{-8}$ & Small constant added to the group standard deviation to prevent division by zero. \\
\midrule
\multicolumn{3}{l}{\textit{Policy Optimization (PPO-style)}} \\
PPO Clipping Parameter ($\epsilon$) & 0.2 & The standard clipping ratio to prevent destructively large policy updates. \\
KL Penalty Coefficient ($\beta_{\text{KL}}$) & 0.01 & Weight of the KL divergence term in Equation~\ref{eq:final_loss} constraining the policy against $\pi_{\text{ref}}$. \\
Step-Level Advantage Normalization & mean-std & Standard PPO/GRPO advantage standardization applied to $\{\hat{A}_t\}$ across the batch before computing the surrogate loss. \\
\bottomrule
\end{tabular}
\end{table*}

\subsection{Experimental Settings}
\label{sec:settings}
GEPO is benchmarked against two classes of methods: (i) prompting-based techniques, including ReAct~\cite{Yao2022-xt} and Reflexion~\cite{Shinn2023-ll}; and (ii) reinforcement learning algorithms, including PPO~\cite{Schulman2017-dh}, RLOO~\cite{Ahmadian2024-fg}, GRPO~\cite{Shao2024-jf}, GiGPO~\cite{Feng2025-oa}, and HGPO~\cite{He2026-qk}. All RL methods are trained with the same backbone (Qwen2.5-1.5B/7B for ALFWorld and WebShop; Qwen2.5-3B/7B for search-augmented QA), batch composition, and training budget, with method-specific hyperparameters tuned per the original papers.

\subsection{Dataset Details}
\label{sec:app_datasets}
All datasets and models utilized in this study are publicly available and were used strictly for academic research, consistent with their intended purposes.

\subsubsection{ALFWorld}
The task in ALFWorld is to parse a high-level natural language goal into a sequence of low-level actions within a simulated household environment. The primary challenges are the long-horizon planning required to decompose the goal and the extreme reward sparsity, as meaningful feedback is provided almost exclusively upon successful completion of the entire multi-step task.

\subsubsection{WebShop}
The task in WebShop~\cite{Yao2022-yg} is to browse a realistic e-commerce website to find and purchase a product that satisfies a given user instruction. This requires the agent to perform a sequence of actions like searching for items, filtering attributes, and navigating through product pages. Key difficulties include a vast state space composed of noisy, high-dimensional HTML observations and a large action space of clickable elements, demanding robust exploration under sparse reward conditions.

\subsubsection{Search-Augmented QA}
To evaluate the multi-step tool calling and information synthesis capabilities of LLM agents, we follow the setup in GiGPO~\cite{Feng2025-oa} and incorporate a variety of search-augmented Question Answering (QA) tasks. These benchmarks are categorized into two types:
(1) \textbf{Single-hop QA}: including NQ~\cite{kwiatkowski-etal-2019-natural}, TriviaQA~\cite{joshi-etal-2017-triviaqa}, and PopQA~\cite{mallen-etal-2023-trust}, which require the agent to retrieve and identify a single piece of factual information.
(2) \textbf{Multi-hop QA}: including HotpotQA~\cite{yang-etal-2018-hotpotqa}, 2WikiMultiHopQA~\cite{ho-etal-2020-constructing}, MuSiQue~\cite{Trivedi2021-sq}, and Bamboogle~\cite{Ofir_Press2022-tx}, which demand more complex reasoning as the agent must iteratively use search tools to link multiple pieces of evidence across different documents. The primary challenge in these tasks lies in the agent's ability to maintain a coherent reasoning chain and accurately formulate search queries to bridge information gaps over multiple turns.

\subsection{Implementation Details and Hyperparameters}
\label{sec:appendix_implementation}
Table~\ref{tab:appendix_gepo_hyperparams} lists the hyperparameters used throughout the paper. Below we add a few implementation notes that do not fit into the table.

To construct a topologically meaningful graph without succumbing to topology bloat, we use a pre-trained Sentence-BERT model. Observations are mapped to the same node if their cosine similarity strictly exceeds the threshold $\delta = 0.9$. This zero-shot semantic filtering is extremely lightweight and prevents minor textual variations from creating spurious nodes.

The transition graph is updated asynchronously as each trajectory rollout finishes, rather than batched at the end of an iteration. Because LLM generation has long-tail latency, the graph updates for early-completing trajectories overlap with the generation time of the slower ones, so by the time the rollout phase ends only the last trajectory is left to merge.

\section{Computational Cost Analysis}
\label{sec:appendix_cost_analysis}

\begin{figure}[htbp]
    \centering
    \includegraphics[width=0.9\columnwidth]{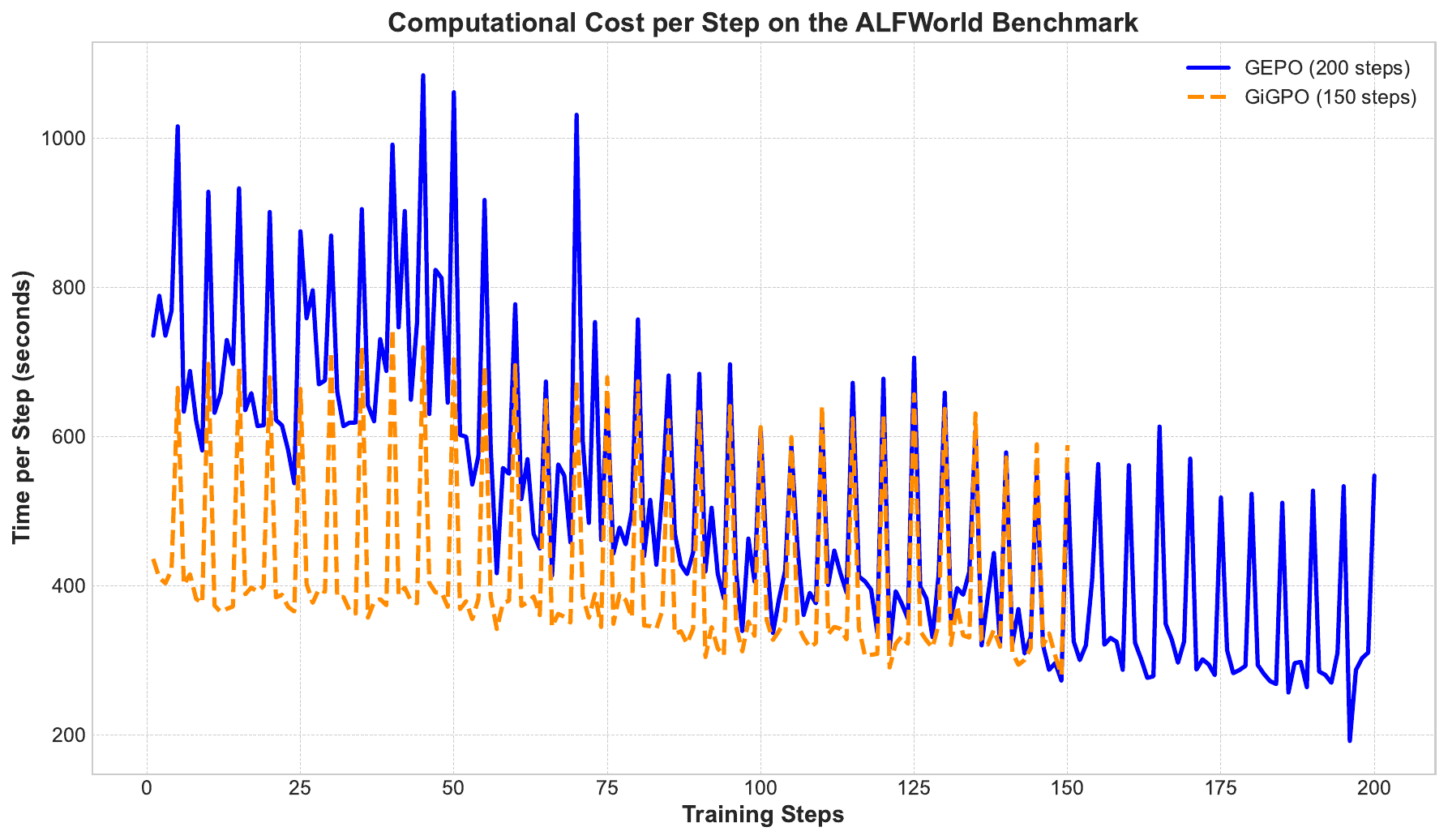}
    \caption{Comparison of computational cost per training step on the ALFWorld benchmark. The solid blue line (GEPO) is more expensive than the dashed orange line (GiGPO), with the gap concentrated in early training (first $\sim$75 steps) and shrinking as the graph stabilizes.}
    \label{fig:appendix_cost}
\end{figure}

We compare the wall-clock training cost of GEPO and the GiGPO baseline in Figure~\ref{fig:appendix_cost}. On average across the entire training process, GEPO incurs a computational overhead of 35\% per step. However, this overhead is temporally asymmetric. During the first $\sim$75 steps, GEPO can be 50--80\% slower. Once the graph stabilizes and a more competent policy produces shorter trajectories, the per-step cost converges, reducing the GEPO overhead to marginal levels in later stages. Overall, GEPO trades a moderate 35\% training-time overhead for the success-rate and stability gains reported in Section~\ref{sec:overall_performance}.
\section{Extension to Vision-Language Environments}
\label{sec:appendix_vlm}

\begin{table}[htbp]
\centering
\caption{Success rates (\%) of Vision-Language Model (VLM) agents using Qwen2.5-VL-3B-Instruct, averaged over 3 seeds. Best results are in \textbf{bold}.}
\label{tab:vlm_results}
\begin{tabular}{llcc}
\toprule
\textbf{Type} & \textbf{Method} & \textbf{Sokoban} & \textbf{EZPoints} \\
\midrule
RL  & GRPO & 67.1 $\pm 4.7$ & 86.9 $\pm 3.4$ \\
RL  & GiGPO$_{\text{w/ std}}$ & 76.9 $\pm 2.7$ & \textbf{100.0 $\pm 0.0$} \\
RL  & GiGPO$_{\text{w/o std}}$ & 81.0 $\pm 3.6$ & \textbf{100.0 $\pm 0.0$} \\
RL  & \textbf{GEPO (Ours)} & \textbf{85.0 $\pm 1.4$} & \textbf{100.0 $\pm 0.0$} \\
\bottomrule
\end{tabular}
\end{table}

To verify GEPO's transferability to Vision-Language Models (VLMs), we evaluate it on Sokoban~\cite{SchraderSokoban2018} and EZPoints~\cite{Zhai2024-ze} using Qwen2.5-VL-3B-Instruct~\cite{Bai2025-rc}. To process visual observations, we replace the Sentence-BERT text encoder with Qwen-3-VL-Embedding-8B~\cite{qwen3vlembedding} and apply the same cosine-similarity merging rule to the resulting multimodal embeddings, leaving the rest of the GEPO pipeline unchanged.

As shown in Table~\ref{tab:vlm_results}, both GEPO and GiGPO saturate EZPoints at 100\% success. On Sokoban, a puzzle environment with irreversible dead-ends, GEPO outperforms the stronger GiGPO variant (w/o std) by 4.0\% with markedly lower variance. The gap is consistent with the analysis in Section~\ref{sec:local_adv_revised}: exact cross-trajectory state matches are rare in visual spaces, leaving horizontal-matching methods like GiGPO vulnerable to visual representation variations, while trajectory-level methods like GRPO fail to isolate early critical moves. By using cross-trajectory matches only to construct the graph topology rather than as the unit of credit attribution, GEPO confines the impact of visual matching noise to second-order changes in graph centrality, demonstrating its generality beyond text-only environments.
\section{Analysis of Topological Centrality Measures}
\label{sec:ablation_centrality}

\begin{table}[htbp]
\centering
\caption{Impact of Different Topological Centrality Measures on ALFWorld Performance across Model Scales. Overall success rates (\%) are averaged over 3 seeds. Best results are in \textbf{bold}.}
\label{tab:centrality_analysis_scaled}
\begin{tabular}{lcc}
\toprule
\textbf{Centrality Metric} & \textbf{1.5B (\%)} & \textbf{7B (\%)} \\
\midrule
\textbf{Betweenness (GEPO)} & \textbf{92.9 $\pm$ 1.1} & \textbf{96.3 $\pm$ 1.2} \\
Eigenvector  & 89.5 $\pm$ 2.4 & 93.8 $\pm$ 2.1 \\
Degree       & 87.2 $\pm$ 3.5 & 91.5 $\pm$ 4.0 \\
Closeness    & 85.8 $\pm$ 3.8 & 89.4 $\pm$ 4.5 \\
\bottomrule
\end{tabular}
\end{table}

We compare topological centrality metrics for the topological prior $C_v$ on ALFWorld (Table~\ref{tab:centrality_analysis_scaled}). Betweenness yields the highest success rate and lowest variance at both model scales. This advantage arises because betweenness naturally isolates nodes lying on shortest paths, which in embodied environments closely align with causal bottlenecks. In contrast, degree centrality captures only local connectivity, eigenvector centrality emphasizes embeddedness in densely interconnected subnetworks, and closeness centrality minimizes average path length to all other nodes---none of which specifically identify the bottleneck transitions that betweenness highlights. Consequently, alternative metrics tend to spread credit across high-traffic but off-path nodes rather than concentrating it on causal bridges, leading to higher variance and suboptimal credit assignment.

\section{Generality of GEPO on the Qwen3 Model Families}
\label{sec:app_qwen3}

\begin{table*}[h]
\centering
\caption{Performance comparison on search-augmented QA tasks using Qwen3-1.7B and Qwen3-8B~\cite{Yang2025-te}. All metrics denote the evaluation success rate (\%).}
\label{tab:qwen3_qa_performance}
\resizebox{\textwidth}{!}{
\begin{tabular}{lllcccccccc}
\toprule
\multirow{2}{*}{\textbf{Model}} & \multirow{2}{*}{\textbf{Type}} & \multirow{2}{*}{\textbf{Method}} & \multicolumn{3}{c}{\textbf{Single-Hop QA}} & \multicolumn{4}{c}{\textbf{Multi-Hop QA}} & \multirow{2}{*}{\textbf{Avg.}} \\
\cmidrule(lr){4-6} \cmidrule(lr){7-10}
& & & \textbf{NQ} & \textbf{TriviaQA} & \textbf{PopQA} & \textbf{HotpotQA} & \textbf{2Wiki} & \textbf{MuSiQue} & \textbf{Bamboogle} & \\
\midrule
\multirow{2}{*}{\textit{Qwen3-1.7B}}
& RL Training & GiGPO & 42.0 & 59.0 & 45.0 & 37.8 & 36.7 & 13.9 & 31.9 & 38.0 \\
& RL Training & \textbf{GEPO (Ours)} & \textbf{44.1} & \textbf{62.2} & \textbf{47.4} & \textbf{40.3} & \textbf{38.8} & \textbf{16.5} & \textbf{36.1} & \textbf{40.8} \\
\midrule
\multirow{2}{*}{\textit{Qwen3-8B}}
& RL Training & GiGPO & 45.1 & 60.3 & 42.1 & 39.0 & 40.9 & 17.7 & 41.8 & 41.0 \\
& RL Training & \textbf{GEPO (Ours)} & \textbf{47.9} & \textbf{63.8} & \textbf{45.2} & \textbf{41.6} & \textbf{43.7} & \textbf{20.9} & \textbf{45.3} & \textbf{44.1} \\
\bottomrule
\end{tabular}
}
\end{table*}

To evaluate generalization, we test GEPO on Qwen3 (1.7B and 8B) using the same search-augmented QA benchmarks as our main experiments. As shown in Table~\ref{tab:qwen3_qa_performance}, GEPO consistently outperforms the strong GiGPO baseline, yielding average improvements of +2.8\% (1.7B) and +3.1\% (8B).

We highlight two key patterns: (1) Gains are most pronounced on tasks requiring longer reasoning chains (e.g., +4.2\% on Bamboogle at 1.7B), validating GEPO's step-level redistribution over extended horizons. (2) The performance gap is consistent across both model sizes, suggesting that GEPO's graph-derived credit signal benefits search-augmented QA independently of the underlying model scale.

Finally, while Qwen3's absolute scores trail Qwen2.5 (aligning with concurrent findings on its agentic capabilities~\cite{Zhao2026-ko}), GEPO's relative advantage over GiGPO remains robust across all evaluated settings.

\begin{table*}[htbp]
\centering
\caption{A side-by-side comparison of agent trajectories in ALFWorld. Both agents are tasked with finding a clean soapbar and placing it in a cabinet. They share an identical history for the first two steps. The divergence at Step 3 highlights GEPO's superior efficiency and non-looping exploration strategy. Reasoning excerpts are verbatim from agent rollouts, abbreviated for space.}
\label{tab:case_study}
\resizebox{\textwidth}{!}{%
\begin{tabular}{p{0.05\linewidth}|p{0.45\linewidth}|p{0.45\linewidth}}
\toprule
\textbf{Step} & \textbf{GEPO (Task: put a clean soapbar in cabinet)} & \textbf{GiGPO (Task: put a clean soapbar in cabinet)} \\
\midrule
\textbf{1} & \textbf{Action:} go to countertop 1 \newline \textbf{Observation:} Arrives at countertop 1, sees a candle 3 and a spraybottle 2. & \textbf{Action:} go to countertop 1 \newline \textbf{Observation:} Arrives at countertop 1, sees a candle 3 and a spraybottle 2. \\
\hline
\textbf{2} & \textbf{Action:} go to toilet 1 \newline \textbf{Observation:} Arrives at toilet 1, sees a toiletpaper 1. & \textbf{Action:} go to toilet 1 \newline \textbf{Observation:} Arrives at toilet 1, sees a toiletpaper 1. \\
\hline
\textbf{3} & \textbf{Reasoning:} \textit{"The task is to put a clean soapbar in the cabinet. I have already explored countertop 1 and toilet 1, and neither contains the target object. Based on the spatial structure, the next logical step is to check unvisited locations where a soapbar might be found. The action 'go to cabinet 1' is chosen as it is a new, promising location..."} \newline\newline \textbf{Action:} go to cabinet 1 & \textbf{Reasoning:} \textit{"The current situation is that you are at toilet 1, and there is no soapbar here. Your task is to put a clean soapbar in a cabinet. You have already taken the first step to go to countertop 1, but you might need to search that area more thoroughly to find the soapbar."} \newline\newline \textbf{Action:} go to countertop 1 \\
\midrule
\textbf{} & GEPO uses the explored portion of the graph to skip locations it has already checked, and moves to a new location (cabinet 1). & GiGPO does not retain this kind of graph-level memory and returns to countertop 1, a location it has already visited. \\
\bottomrule
\end{tabular}%
}
\end{table*}
\section{Qualitative Case Study in ALFWorld}
\label{sec:appendix_case_study}

To qualitatively demonstrate how GEPO mitigates the structural blindness that affects standard group-based RL agents, we present a comparative case study from the ALFWorld environment. We analyze the trajectories of two agents---one trained with GEPO and a strong GiGPO baseline—assigned the same task and starting with the same initial exploration history. The task is to locate an item and place it in a cabinet. As shown in Table~\ref{tab:case_study}, both agents first explore countertop 1 and then toilet 1, finding neither location fruitful. The critical divergence occurs at Step 3.

Although the GiGPO agent acknowledges in its reasoning that countertop 1 has already been checked, it returns to that location at step 3. The agent does not carry a representation of the explored portion of the environment across reasoning steps, so the cue does not translate into a different action. The GEPO agent concludes that since the countertop and toilet have been inspected, it should try a new location and selects the cabinet. This is consistent with the criticality signal it was trained against: states that have already been visited contribute little to the redistributed advantage, while unvisited states with high task-conditioned criticality are favored.


\section{Prompts}

\begin{figure*}[htbp] 
    \centering
    \begin{promptbox}{Prompt Template for ALFWorld}
      You are an expert agent operating in the ALFWorld embodied environment. Your task is to: \var{task\_description}. Prior to this step, you have already taken \var{step\_count} step(s). Below are the most recent \var{history\_length} observations and the corresponding actions you took: \var{action\_history}. You are now at step \var{current\_step} and your current observation is: \var{current\_observation}. Your admissible actions of the current situation are: [\var{admissible\_actions}].

Now it's your turn to take an action. You should first reason step-by-step about the current situation. This reasoning process MUST be enclosed within \textbf{\texttt{<think>}} and \textbf{\texttt{</think>}} tags. Once you've finished your reasoning, you should choose an admissible action for current step and present it within \textbf{\texttt{<action>}} and \textbf{\texttt{</action>}} tags.
    \end{promptbox}
    \caption{The prompt template of ALFWorld agents.} 
    \label{fig:alfworld_prompt}
\end{figure*}

\begin{figure*}[htbp]
    \centering
    \begin{promptbox}{Prompt Template for WebShop}
      You are an expert autonomous agent operating in the WebShop e-commerce environment. Your task is to: \var{task\_description}. Prior to this step, you have already taken \var{step\_count} step(s). Below are the most recent \var{history\_length} observations and the corresponding actions you took: \var{action\_history}. You are now at step \var{current\_step} and your current observation is: \var{current\_observation}. Your admissible actions for the current situation are: [\var{available\_actions}].
      
      \smallskip
      Now it's your turn to take one action for the current step. You should first reason step-by-step about the current situation, then think carefully which admissible action best advances the shopping goal. This reasoning process MUST be enclosed within \textbf{\texttt{<think>}} and \textbf{\texttt{</think>}} tags. Once you've finished your reasoning, you should choose an admissible action for current step and present it within \textbf{\texttt{<action>}} and \textbf{\texttt{</action>}} tags.
    \end{promptbox}
    \caption{The prompt template used for WebShop agents.}
    \label{fig:webshop_prompt}
\end{figure*}

\begin{figure*}[htbp]
    \centering
    \begin{promptbox}{Prompt Template for Search}
      You are an expert agent tasked with answering the given question step-by-step. Your question: \var{task\_description}. Prior to this step, you have already taken \var{step\_count} step(s). Below is the interaction history where \textbf{\texttt{<search>}} \textbf{\texttt{</search>}} wrapped your past search queries and \textbf{\texttt{<information>}} \textbf{\texttt{</information>}} wrapped the corresponding search results returned by the external search engine. History: \var{memory\_context}
      
      \smallskip
      Now it's your turn to respond for the current step. You should first conduct reasoning process. This process MUST be enclosed within \textbf{\texttt{<think>}} and \textbf{\texttt{</think>}} tags. After completing your reasoning, choose only one of the following actions (do not perform both):
      
      (1) If you find you lack some knowledge, you can call a search engine to get more external information using format: \textbf{\texttt{<search>}} your query \textbf{\texttt{</search>}}.
      
      (2) If you have enough knowledge to answer the question confidently, provide your final answer within \textbf{\texttt{<answer>}} \textbf{\texttt{</answer>}} tags, without detailed illustrations. For example, \textbf{\texttt{<answer>}}Beijing\textbf{\texttt{</answer>}}.
    \end{promptbox}
    \caption{The prompt template of Search agents.}
    \label{fig:search_prompt}
\end{figure*}

\end{document}